\documentclass{article}
\PassOptionsToPackage{numbers, compress}{natbib}

\usepackage[final]{neurips_2020}

\usepackage[utf8]{inputenc} 
\usepackage[T1]{fontenc}    
\usepackage{hyperref}       
\usepackage{url}            
\usepackage{booktabs}       
\usepackage{amsfonts}       
\usepackage{nicefrac}       
\usepackage{microtype}      
\usepackage{latexsym,amscd,amsbsy}
\usepackage{mathtools}
\usepackage{xcolor}
\usepackage{enumitem}

\def\DD{\text{D}}

\def\dd{\text{d}}

\def\sign{\text{sign}}
\def\new{\text{new}}
\def\erf{\text{erf}}

\def\erfc{\text{erfc}}

\let\a=\alpha   
   
\let\l=\lambda

\let\D=\Delta \let\L=\Lambda

\def\arccosh{{\rm arccosh}}

\def\DD{{\cal D}} 
\def\KK{{\cal K}}  
\def\ZZ{{\cal Z}}

\def \bfw{{\bf w}}

  \def\erf{\text{erf}}

\def\de{\mathrm d}

\def\to{\rightarrow}

\newcommand{\beq}{\begin{equation}} \newcommand{\eeq}{\end{equation}}

\title{Dynamical mean-field theory for stochastic gradient descent in Gaussian mixture classification}

\author{
Francesca Mignacco$^1$\And Florent Krzakala$^{2,3}$ \And Pierfrancesco Urbani$^1$\And Lenka Zdeborov\'a$^{1,4}$\\\\
$^1$ Institut  de  physique th\'eorique,  Universit\'e  Paris-Saclay,  CNRS,  CEA,  Gif-sur-Yvette,  France\\
$^2$ Laboratoire de Physique, CNRS, \'Ecole Normale Sup\'erieure, PSL University, Paris, France\\
$^3$ IdePHICS Laboratory, EPFL, Switzerland\\
$^4$ SPOC Laboratory, EPFL, Switzerland\\
Correspondence to: \texttt{francesca.mignacco@ipht.fr}
}

\begin{document}
\maketitle

\begin{abstract}
We analyze in a closed form the learning dynamics
of stochastic gradient descent (SGD) for a single-layer neural
network classifying a high-dimensional Gaussian mixture where each cluster is assigned one of two labels. 
This problem provides a prototype of a non-convex loss landscape with interpolating regimes and a large generalization gap.   
We define a particular stochastic process for
which SGD can be extended to a continuous-time limit that we call
stochastic gradient flow. 
In the full-batch limit, we recover the standard gradient flow.
We apply dynamical mean-field theory from statistical physics to
track the dynamics of the algorithm in the high-dimensional limit via
a self-consistent stochastic process. 
We explore the performance of the algorithm as a function of the control
parameters shedding light on how it navigates the loss landscape.
\end{abstract}

\section{Introduction}

Understanding how stochastic gradient descent (SGD) manages to train
artificial neural networks with good generalization capabilities by
exploring the high-dimensional non-convex loss landscape is one of the
central problems in the theory of machine learning.  A popular attempt to
explain this behavior is by showing that the loss landscape itself is
simple, with no spurious (i.e. leading to bad test error) local
minima. Some empirical evidence instead leads to the conclusion that the loss
landscape of state-of-the-art deep neural networks actually has
spurious local (or even global) minima and stochastic gradient descent
is able to find them \cite{safran2017spurious,LPA19}. Still, the
stochastic gradient descent algorithm, initialized at random, leads to
good generalization properties in practice. It became clear that a
theory that would explain this success needs to account for the whole
trajectory of the algorithm. Yet this remains a challenging task,
certainly for the state-of-the art deep networks trained on real
datasets.

\paragraph{Related work ---} A detailed description of the whole
trajectory taken by the (stochastic) gradient descent was so far
obtained only in several special cases. First such case are deep
linear networks where the dynamics of gradient descent has been
analyzed \cite{BO97,SMG13}. While this line of works has led to very
interesting insights about the dynamics, linear networks lack the
expressivity of the non-linear ones and the large-time behavior of the
algorithm can be obtained with a simple spectral algorithm. Moreover,
the analysis of dynamics in deep linear networks was not extended to
the case of stochastic gradient descent. Second case where the
trajectory of the algorithm was understood in detail is the
\emph{one-pass} (online) stochastic gradient descent for two-layer
neural networks with a small hidden layer in the teacher-student
setting \cite{SS95Short, SS95,Sa09,
  GASKZ19,goldt2019modelling}. However, the {one-pass} assumption made
in those analyses is far from what is done in practice and is unable
to access the subtle difference between the training and test error
that leads to many of the empirical mysteries observed in deep
learning. A third very interesting line of research that recently
provided insight about the behavior of stochastic gradient descent
concerns two-layer networks with divergingly wide hidden layer. This
mean-field limit \cite{rotskoff2018neural,MMN18,chizat2018global} maps
the dynamics into the space of functions where its description is
simpler and the dynamics can be written in terms of a closed set of
differential equations. It is not clear yet
whether this analysis can be extended in a sufficiently explicit way
to deeper or finite width neural networks. The term \textit{mean-field} has been used in several contexts in machine learning \cite{NIPS2016_6322,schoenholz2017deep,yang2018a,mei2019meanfield,gilboa2019dynamical,novak2019bayesian}. Note that the term in the aforementioned works refers to a variety of approximations and concepts. In this work we use it with the same meaning as in \cite{MPV87,GKKR96, PUZ20}. Most importantly, the term mean-field in our case has nothing to do with the width of an eventual hidden layer. We refer to \cite{Gabri__2020} for a broader methodological review of mean-field methods and their applications to neural networks.

Our present work inscribes in the above line of research offering the
dynamical mean-field theory (DMFT) formalism \cite{MPV87,GKKR96, PUZ20} leading
to a closed set of integro-differential equations to track the full
trajectory of the gradient descent (stochastic or not) from random
initial condition in the high-dimensional limit for in-general
non-convex losses. While in general the DMFT is a heuristic
statistical physics method, it has been amenable to rigorous proof in
some cases \cite{arous1997symmetric}. This is hence an important future
direction for the case considered in the present paper. The DMFT has
been applied recently to a high-dimensional inference problem in
\cite{SBCKUZ20,SKUZ19} studying the spiked matrix-tensor
model. However, this problem does not allow a natural way to study the
stochastic gradient descent or to explore the difference between
training and test errors. In particular, the spiked matrix-tensor
model does not allow for the study of the so-called interpolating
regime, where the loss function is optimized to zero while the test
error remains positive. As such, its landscape is intrinsically
different from supervised learning problems since in the former the
spurious minima proliferate at high values of the loss while the good
ones lie at the bottom of the landscape. Instead, deep networks have
both spurious and good minima at 100\% training accuracy and their
landscape resembles much more the one of continuous constraint
satisfaction problems \cite{FPSUZ17, FHU19}.

\paragraph{Main contributions ---}  We study a natural problem of supervised
classification where the input data come from a high-dimensional
Gaussian mixture of several clusters, and all samples in one cluster are assigned to one of two possible output labels. We then consider a
single-layer neural network classifier with a general non-convex loss function. We analyze a stochastic gradient descent algorithm in which, at each iteration, the batch used to compute the gradient of the loss is 
extracted at random, and we define a particular stochastic process for which SGD can be extended to a continuous-time limit that we call stochastic gradient flow (SGF). 
In the full-batch limit we recover the standard Gradient Flow (GF). 
We describe the high-dimensional limit of the randomly initialized SGF with the DMFT that leads to a description of the dynamics in terms of a self-consistent stochastic process that we compare with numerical simulations. In particular, we show that the finite batch size can have a beneficial effect in
the test error and acts as an effective regularization that prevents overfitting.

\section{Setting and definitions}
\label{sec:two_cluster} \label{sec:three_cluster}

In all what follows, we will consider the high-dimensional setting where the dimension of each point in the dataset is $d\to \infty$ and the size of the training set $n=\alpha d$, being $\alpha$ a control parameter that we keep of order one.

We consider a training set made of $n$ points 
\beq
{\bf X}=({\bf x}_1, ... {\bf x}_n)^\top\in \mathbb{R}^{n\times d}\ \ \ \ \textrm{with labels}\ \ \ \  {\bf y}=(y_1,...y_n)^\top\in\{+1,-1\}^n.
\eeq
The patterns ${\bf x}_{\mu}$ are given by
\beq
{\bf x}_{\mu}=c_{\mu}\frac{{\bf
    v^\ast}}{\sqrt{d}}+\sqrt{\Delta}\,{\bf z}_{\mu},\quad \quad 
{\bf z}_{\mu}\sim \mathcal{N}({\bf 0}, {\bf I}_d),\ \ \ \ \ \mu=1,...n\:.
\eeq 
Without loss of generality, we choose a basis where  ${\bf v}^\ast=(1,1,...1)\in \mathbb{R}^d$.

\paragraph{Two-cluster dataset:} 
We will illustrate our results on a two-cluster example where the coefficients $c_\mu$ are
taken at random  $c_\mu=\pm 1$ with equal probability.
Therefore one has two symmetric clouds of Gaussian points centered around two vectors ${\bf v}^\ast$ and $-{\bf v}^\ast$.
The labels of the data points are fixed by $y_\mu= c_\mu$.
If the noise level $\Delta$ of the number of samples is small enough,
the two Gaussian clouds are linearly separable by an hyperplane, as
specified in detail in \cite{mignacco2020role},  
and therefore a single layer neural network is enough to perform the
classification task in this case.
We hence consider learning with the simplest neural network that classifies the data according to
$
\hat y_\mu({\bf w})=\textrm{sgn} [ {{\bf w}^\top {\bf x}_\mu}/{\sqrt
  d} ]
$.

\paragraph{Three-cluster dataset:}  We consider also an example of three clusters where a good generalization error
cannot be obtained by separating the points linearly. 
In this case we define $c_\mu =0$ with probability $1/2$, and $c_\mu
=\pm 1$ with probability $1/2$. 
The labels are then assigned as
\beq
y_\mu=-1  {~\rm if~}  c_\mu=0 \text{~~, and~~} y_\mu=1    {~\rm if~~}
c_\mu=\pm 1
\, .
\label{ynl}
\eeq 
One has hence three clouds of Gaussian points, two
external and one centered in zero.  
In order to fit the data we consider a single layer-neural network
with the door activation function, defined as
\beq
\hat y_\mu({\bf w}) =\textrm{sgn}\left[\left(\frac {{\bf w}^\top {\bf x}_\mu}{\sqrt d}\right)^2-L^2\right].
\eeq
The onset parameter $L$ could be learned, but we will instead fix it to a constant.

\paragraph{Loss function:} \label{sec:loss}
We study the dynamics of learning by
the empirical risk minimization of the loss 
\begin{equation}
\mathcal{H}({\bf w}) = \sum_{\mu=1}^n \ell \left[ y_{\mu}\phi\left(\frac{{\bf w}^\top {\bf x}_{\mu}}{\sqrt{d}}\right)\right]+\frac{\lambda}{2}\Vert{\bf w}\Vert_2 ^2,
\label{loss}
\end{equation}
where we have added a Ridge regularization term.
The activation function $\phi$ is given by
\beq
\phi(x) =\begin{cases}
x & \textrm{linear for the two-cluster dataset}\\
x^2-L^2  & \textrm{door for the three-cluster dataset}\:.\label{Phi_def}
\end{cases}
\eeq
The DMFT analysis is valid for a generic loss function $\ell$. 
However, for concreteness, in the result section we will focus on the logistic loss
$
\ell(v) = \ln \left(1+e^{-v}\right)\:
$.
Note that in this setting the two-cluster dataset leads to convex
optimization, with a unique minimum for finite $\l$, and implicit
regularization for $\l=0$ \cite{rosset2004margin}, and was analyzed
in detail in \cite{deng2019model,mignacco2020role}. Still the
performance of stochastic gradient descent with finite batch size
cannot be obtained in \emph{static} ways. 
The three-cluster dataset,
instead, leads to a generically non-convex optimization problem which
can present many spurious minima with different generalization abilities when the control parameters such as $\D$ and $\a$ are changed.
We note that our analysis can be extended to neural networks with a small hidden layer
\cite{SOS92}. This would allow to study the role of overparametrization, but it is left for future work. 

\section{Stochastic gradient-descent training dynamics}
\label{sec:SGF}
\paragraph{Discrete SGD dynamics ---} We consider the discrete gradient-descent dynamics for which the weight update is given by
\beq
{\rm w}_j(t+\eta)={\rm w}_j(t) -\eta\left[\lambda {\rm w}_j(t) +\sum_{\mu=1}^n \,s_{\mu}(t) \L'\left(y_{\mu} ,\frac{{\bf w}(t)^\top {\bf x}_{\mu}}{\sqrt{d}}\right)  \frac{{\rm x}_{\mu,j}}{\sqrt{d}}\right]\label{SGDdiscrete_dynamics}
\eeq
where we have introduced the function
$
\L(y,h) = \ell\left(y\phi\left(h\right)\right)
$
and we have indicated with a prime the derivative with respect to $h$, i.e.,
$
\Lambda'(y, h) = y\ell'\left(y\phi\left(h\right)\right) \phi'\left(h\right)
$. We consider the following initialization of the weight vector
$
{\bf w}(0)\sim \mathcal{N}({\bf 0}, {\bf I}_d R)
$, where $R>0$ is a parameter that tunes the average length of the
weight vector at the beginning of the dynamics\footnote{The DMFT equations we derive can be easily generalized to the case in
which the initial distribution over $\bf w$ is different. We only
need it to be separable and independent of the dataset. }. 
The variables $s_\mu(t)$ are i.i.d. binary random
variables. Their discrete-time dynamics can be chosen in two ways:
\begin{itemize}[leftmargin=*,nosep]
\item In classical {\bf SGD}, when sampling {\em with replacement}, at iteration $t$ one extracts the samples with the following probability distribution
\beq
s_\mu(t) =\begin{cases}
1 & \textrm{with probability}\ \ \ b\\
0 & \textrm{with probability}\ \ \ 1-b\\
\end{cases}
\label{real_SGD}
\eeq
and $b\in (0,1]$. In this way for each time iteration one extracts on average $B=bn$ patterns at random on which the gradient is computed and therefore the batch size is given by $B$. Note that if $b=1$ one recovers full-batch gradient descent.

\item {\bf Persistent SGD} is defined by a stochastic process for $s_\mu(t)$ given by the following probability rules
\beq
\begin{split}
&{\rm Prob}(s_\mu(t+\eta)=1|s_\mu(t)=0) =\frac{1}{\tau}\eta\\
&{\rm Prob}(s_\mu(t+\eta)=0|s_\mu(t)=1) = \frac{(1-b)}{b\tau}\eta ,
\end{split}
\label{PSGD}
\eeq
where $s_{\mu}(0)$ is drawn from the probability distribution \eqref{real_SGD}. In this case, for each time slice one has on average $B=bn$ patterns that are \emph{active}  and enter in the computation of the gradient. The main difference with respect to the usual SGD is that one keeps the same patterns and the same minibatch for a characteristic time $\tau b /(1-b)$. Again, setting $b=1$ one gets full-batch gradient descent and all the patterns are always active.
\end{itemize}

\paragraph{Stochastic gradient flow ---} To write the DMFT we consider
a continuous-time dynamics defined by the $\eta\to 0$ limit. 
This limit is not well defined for the usual SGD dynamics
described by the rule \eqref{real_SGD} and we consider instead its \emph{persistent} version described by Eq.~\eqref{PSGD}. In this case the stochastic process for $s_\mu(t)$ is well defined for $\eta\to 0$ and one can write a continuous-time equation as
\begin{equation}
\dot{\rm w}_j(t)= -\lambda {\rm w}_j(t) -\sum_{\mu=1}^n\,s_{\mu}(t) \L'\left( y_{\mu}, \frac{{\bf w}(t)^\top {\bf x}_{\mu}}{\sqrt{d}}\right)\frac{{\rm x}_{\mu,j}}{\sqrt{d}},\label{SGD_flow}
\end{equation}
Again, for $b=1$ one recovers the gradient flow. We call Eq.~(\ref{SGD_flow}) {\em stochastic gradient flow} (SGF).

\section{Dynamical mean-field theory for SGF}
\label{DMFTforSGF}
We will now analyze the SGF in the infinite size limit $n\to \infty$, $d \to \infty$ with $\a=n/d$ and $b$ and $\tau$ fixed and of order one. 
In order to do that, we use dynamical mean-field theory (DMFT).
The derivation of the DMFT equations is given in the supplementary material, here we will just present the main steps.
The derivation extends the one reported in \cite{ABUZ18} for the
non-convex perceptron model \cite{FPSUZ17} (motivated there as a model of
glassy phases of hard spheres). The main differences of the present
work with respect to \cite{ABUZ18} are that here we consider a finite-batch gradient descent and that our dataset is structured while in
\cite{ABUZ18} the derivation was done for full-batch gradient descent
and random i.i.d. inputs and i.i.d. labels, i.e. a case where one cannot
investigate generalization error and its properties.  The starting point of the DMFT is the dynamical partition function
\begin{equation}
Z_{\rm dyn}= \int_{{\bf w}(0)={\bf w}^{(0)}} \!\!\! \mathcal{D}{\bf w}(t)\prod_{j=1}^d \,\delta\left[-\dot {\rm w}_j(t)-\lambda {\rm w}_j(t) -\sum_{\mu=1}^n \,s_{\mu}(t) \L'\left(y_{\mu}, \frac{{\bf w}(t)^\top {\bf x}_{\mu}}{\sqrt{d}}\right)\frac{{\rm x}_{\mu,j}}{\sqrt{d}}\right] ,
\label{Zdyn}
\end{equation}
where $\DD {\bf w}(t)$ stands for the measure over the dynamical
trajectories starting from ${\bf w}(0)$. Since $Z_{\rm dyn}=1$ (it is
just an integral of a Dirac delta function) \cite{De76} one can average directly $Z_{\rm dyn}$ over the training set, the initial condition and the stochastic processes of $s_\mu(t)$. We indicate this average with the brackets $\langle\cdot\rangle$. Hence we can write
\begin{equation}
\begin{split}
Z_{\rm dyn}= \left\langle\int \mathcal{D}{\bf w}(t)\mathcal{D}\hat {\bf w}(t) \,e^{S_{\rm dyn}}\right\rangle,
\end{split}
\end{equation} 
where we have defined 
\begin{equation}
S_{\rm dyn}=\sum_{j=1}^d \int_0^{+\infty}\dd t \,i{\hat{\rm w}_j(t)}\left( -\dot{\rm w}_j(t)-\lambda {\rm w}_j(t) -\sum_{\mu=1}^n \,s_{\mu}(t) \L'\left( y_{\mu}, \frac{{\bf w}(t)^\top {\bf x}_{\mu}}{\sqrt{d}}\right)\frac{{\rm x}_{\mu,j}}{\sqrt{d}}\right).\label{Sdyn}
\end{equation}
and we have introduced a set of fields $\hat{\bfw}(t)$ to produce the integral representation of the Dirac delta function. The average over the training set can be then performed explicitly, and the dynamical partition function $Z_{\rm dyn}$ is expressed as an integral of an exponential with extensive exponent in $d$:
\beq
Z_{\rm dyn} = \int \mathcal{D}{\bf Q}\,\mathcal{D}{\bf m}\,\, e^{d S({\bf Q},{\bf m})},
\label{SP_Z}
\eeq
where ${\bf Q}$ and ${\bf m}$ are two dynamical order parameters defined in the supplementary material. Therefore, the dynamics in the $d\to \infty$ limit satisfies a large deviation principle and we can approximate $Z_{\rm dyn}$ with its value at the saddle point of the action $S$. In particular, one can show that the saddle point equations for the parameters ${\bf Q}$ and ${\bf m}$ can be recast into a self consistent stochastic process for a variable $h(t)$ related to the typical behavior of ${{\bf w}(t)^\top {\bf z}_{\mu}}/{\sqrt{d}}$, which evolves according to the stochastic equation:
\beq
\partial_t{h}(t)=-(\l +\hat {\l}(t)) h(t)- \sqrt \D \,s(t)\,\L'\left(y(c),r(t) - Y(t)\right)+\int _{0}^t \dd t' M_R(t,t')h(t')+\xi(t),
\label{ht}
\eeq
where we have denoted by $r(t)=\sqrt \D h(t)+ m(t)(c+\sqrt{\Delta}h_0 )$ and $m(t)$ is the \emph{magnetization}, namely $m(t)={\bf w}(t)^\top {\bf v}^*/d$. The details of the computation are provided in the supplementary material.
There are several sources of stochasticity in Eq.~\eqref{ht}. First, one has a dynamical noise $\xi(t)$ that is Gaussian distributed and characterized by the correlations
\beq
\langle \xi(t)\rangle=0, \ \ \ \ \ \ \ \langle \xi(t)\xi(t')\rangle = M_C(t,t')\:.
\eeq
Furthermore, the starting point $h(0)$ of the stochastic process is random and distributed according to
\beq
P(h(0))=e^{- h(0)^2/(2R)}/\sqrt{2\pi R}\,.
\eeq 
Moreover, one has to introduce a quenched Gaussian random variable $h_0$ with mean zero and average one.
We recall that the random variable $c=\pm1$ with equal probability in the two-cluster model, while $c=0, \pm 1$ in the three-cluster
one. The variable $y(c)$ is therefore $y(c)=c$ in the two-cluster case, and is given by Eq.~\eqref{ynl} in the three-cluster one. Finally, one has a dynamical stochastic process $s(t)$ whose statistical properties are specified in Eq.~\eqref{PSGD}.
The magnetization $m(t)$ is obtained from the following deterministic differential equation 
\beq
\partial_t m (t) = -\lambda {m}(t)-\mu(t), \ \ \ \ \ \ \ \ \ \  \ \ m(0)=0^+\:.
\label{mag_eq}
\eeq
The stochastic process for $h(t)$, the evolution of $m(t)$, as well as the statistical properties of the dynamical noise $\xi(t)$ depend on a series of kernels that must be computed self-consistently and are given by
\beq
\begin{split}
\hat {\l}(t)&=\alpha \Delta\left<s(t)\L'' \left(y(c), r(t)\right)\right>,\\
\mu(t)&=\alpha \left< s(t) \left(c+\sqrt \D h_0\right)\L' \left(y(c), r(t)\right)\right>,\\
M_C(t,t')&=\alpha\D\left< s(t)s(t')\L' \left(y(c),r(t)\right)\L' \left(y(c),r(t')\right)\right>,
\\
M_R(t,t')&= \left.\a  \D \frac{\delta }{\delta Y(t')} \langle s(t) \L'(y(c),r(t))\rangle\right|_{Y=0}.
\end{split}
\label{kernels}
\eeq
In Eq.~\eqref{kernels} the brackets denote the average over all the sources of stochasticity in the self-consistent stochastic process.
Therefore one needs to solve the stochastic process in a self-consistent way.
Note that $Y(t)$ in Eq.~\eqref{ht} is set to zero and we need it only to define the kernel $M_R(t,t')$.
The set of Eqs.~\eqref{ht}, \eqref{mag_eq} and \eqref{kernels} can be solved by a simple straightforward iterative algorithm. 
One starts with a guess for the kernels and then runs the stochastic process for $h(t)$ several times to update the kernels.
The iteration is stopped when a desired precision on the kernels is
reached \cite{EO92}. 

Note that, in order to solve Eqs.~\eqref{ht}, \eqref{mag_eq} and
\eqref{kernels}, one needs to discretize time. In the result section
\ref{result_section}, in order to compare our theoretical predictions with numerical simulations, we will take the
time discretization of DMFT equal to the learning rate in the
simulations. In the time-discretized DMFT, this allows us to extract 
 the variables $s(t)$  either from \eqref{real_SGD} (SGD) or
 \eqref{PSGD} (Persistent SGD). In the former case this provides an SGD-inspired discretization of the DMFT equations, which is exact also in discrete time provided that the weight increments do not have higher-order terms than $\mathcal{O}(\eta)$.

Finally, once the self-consistent stochastic process is solved, one has access also to the dynamical correlation function $
C(t,t') = \bfw(t)\cdot \bfw(t')/d
$, encoded in the dynamical order parameter ${\bf Q}$ that appears in the large deviation principle of Eq.~\eqref{SP_Z}.  The correlation $
C(t,t')$ concentrates for $d\to \infty$ and therefore is controlled by the equations
\beq
\begin{split}
\partial_t C(t',t)=&-\tilde \l (t) C(t,t')+\int_0^t \dd s\, M_R(t,s) C(t',s)+\int_0^{t'}\dd s \,M_C(t,s)R(t',s)\\
&-m(t')\left(\int_0^t \dd s M_R(t,s)m(s)+\mu(t)-\hat{\lambda}(t)m(t)\right) \quad \text{ if } t\neq t',\\
\frac{1}{2}\,\partial_t C(t,t)=&-\tilde \l (t) C(t,t)+\int_0^t \dd s\, M_R(t,s) C(t,s)+\int_0^{t}\dd s \,M_C(t,s)R(t,s)\\
&-m(t)\left(\int_0^t \dd s \,M_R(t,s)m(s)+\mu(t)-\hat{\lambda}(t)m(t)\right), \\
\partial_t R(t,t')&=-\tilde \l(t)R(t,t')+\delta(t-t')+ \int_{t'}^{t} \dd s \, M_R(t,s)R(s,t'),\\
\end{split}
\label{Dyson}
\eeq
where we have used the shorthand notation $\tilde \l(t) = \l+\hat{\l}(t)$. We consider the linear response regime, and $R(t,t')=\sum_i \delta {\rm w}_i(t)/\delta H_i(t')/d$ is a response function that controls the variations of the weights when their dynamical evolution is affected by an infinitesimal local field $H_i(t)$. Coupling a local field $H_i(t)$ to each variable $w_i(t)$ changes the loss function as follows: $\mathcal{H}\left({\bf w}(t)\right)\rightarrow \mathcal{H}\left({\bf w}(t)\right)-\sum_{i=1}^d H_i (t)w_i (t)$, resulting in an extra term $H_i(t)$ to the right hand side of Eq. \eqref{SGD_flow}. We then consider the limit $H_i(t)\rightarrow 0$.
It is interesting to note that the second of Eqs.~\eqref{Dyson} controls the evolution of the norm of the weight vector $C(t,t)$ and even if we set $\l=0$ we get that it contains an effective regularization $\hat \l(t)$ that is dynamically self-generated \cite{SHNGS18}.

\paragraph{Dynamics of the loss and the generalization error ---} Once the solution for the self-consistent stochastic process is found, one can get several interesting quantities.
First, one can look at the training loss, which can be obtained as
\beq
e(t) = \alpha\langle \L( y,r(t))\rangle ,
\eeq
where again the brackets denote the average over the realization of the stochastic process in Eq.~\eqref{ht}.
The training accuracy is given by
\beq
a(t) = 1-\langle \theta(- y \phi(r(t)))\rangle
\eeq
and, by definition, it is equal to one as soon as all vectors in the training set are correctly classified.
Finally, one can compute the generalization error.
At any time step, it is defined as the fraction of mislabeled instances:
\begin{equation}
\varepsilon_{\rm gen}(t)=\frac{1}{4}\mathbb{E}_{{\bf X},{\bf y},{\bf x}_{\rm new},y_{\rm new}}\left[\left(y_{\rm new}-\hat{y}_{\rm new}\left({\bf w}(t)\right)\right)^2\right],
\end{equation}
where $\{{\bf X},{\bf y}\}$ is the training set, ${\bf x}_{\rm new}$ is an unseen data point and $\hat{y}_{\rm new}$ is the estimator for the new label $y_{\rm new}$. The dependence on the training set here is hidden in the weight vector ${\bf w}(t)={\bf w}(t,{\bf X}, {\bf y})$.
In the two-cluster case one can easily show that
\begin{equation}
\varepsilon_{\rm gen}(t)=\frac{1}{2}\erfc\left(\frac{m(t)}{\sqrt{2\Delta \, C(t,t)}}\right).\label{egen_sign}
\end{equation}
Conversely, for the door activation trained on the three-cluster dataset we obtain
\begin{equation}
\varepsilon_{\rm gen}(t)=\frac{1}{2} \erfc\left(\frac{L}{\sqrt{2\Delta C(t,t)}}\right)+\frac{1}{4}\left(\erf\left(\frac{L-m(t)}{\sqrt{2\Delta C(t,t)}}\right)+\erf\left(\frac{L+m(t)}{\sqrt{2\Delta C(t,t)}}\right)\right).
\end{equation}

\section{Results}
\label{result_section}
In this section, we compare the theoretical curves resulting from the solution of the DMFT equations derived in Sec. \ref{DMFTforSGF} to numerical simulations. This analysis allows to gain insight into the learning dynamics of stochastic gradient descent and its dependence on the various control parameters in the two models under consideration. 

\begin{figure}[t]
\includegraphics[scale=0.2]{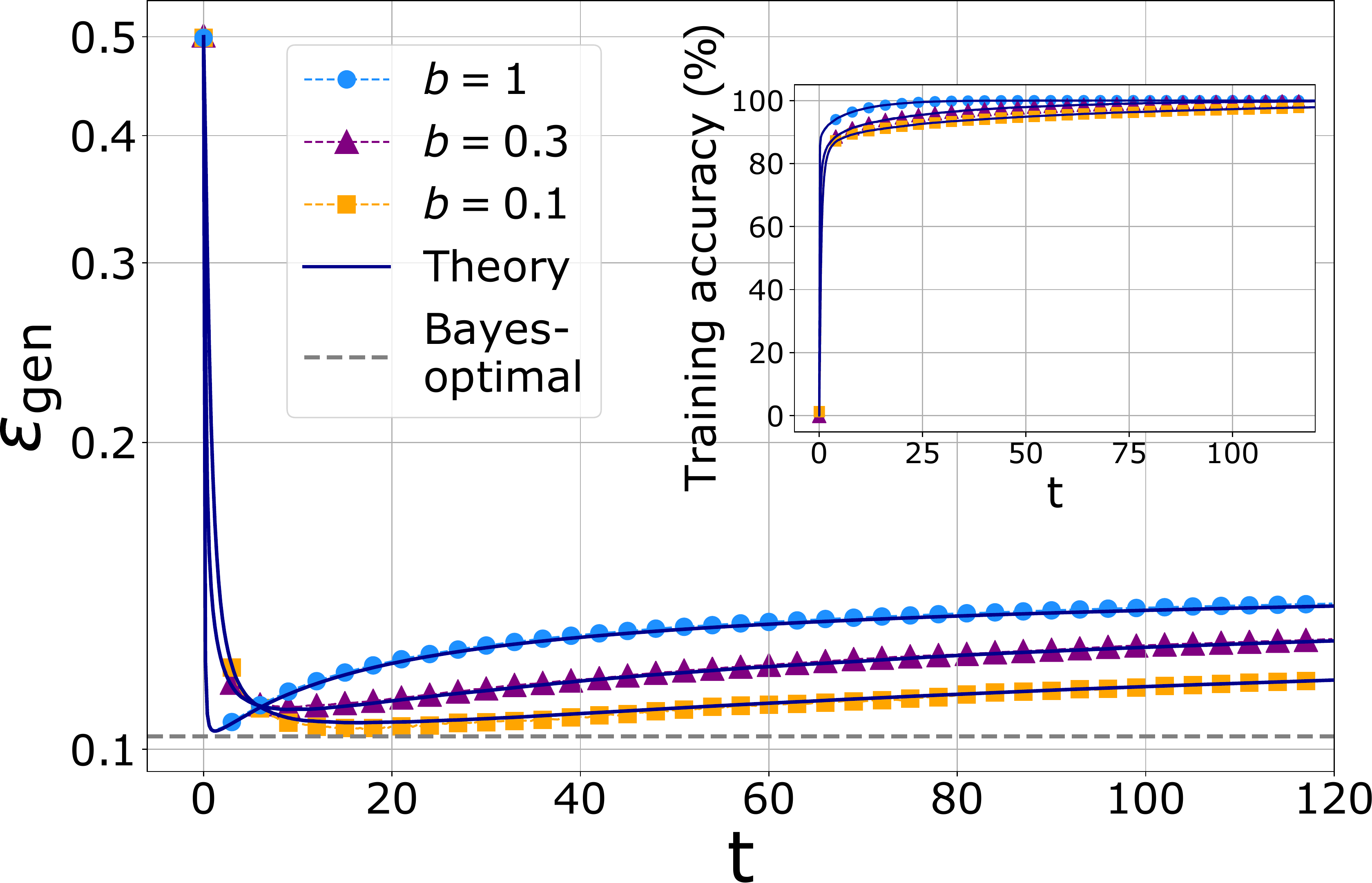} 
\includegraphics[scale=0.2]{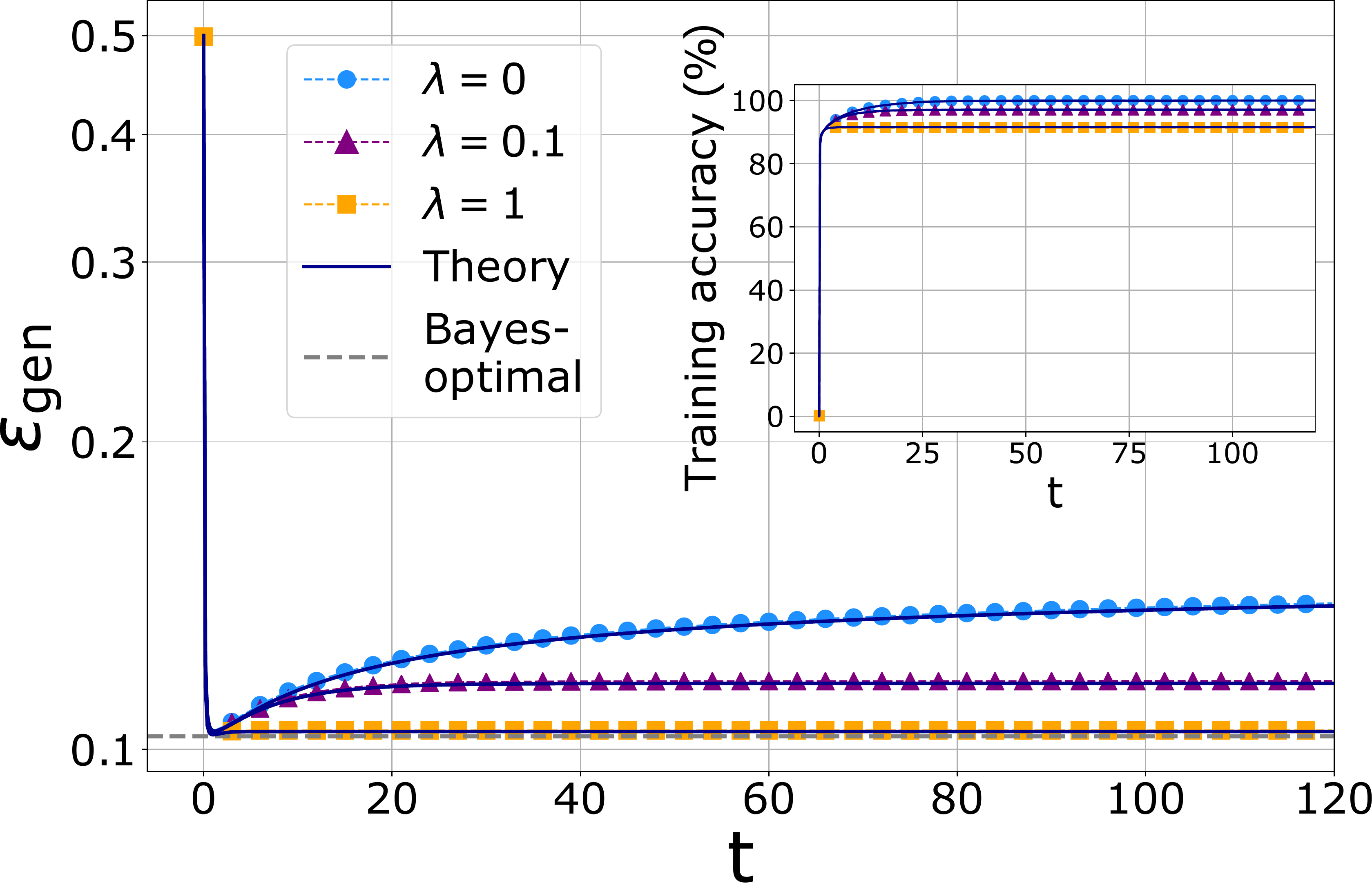} 
\caption{\textbf{Left:} Generalization error as a function of the
  training time for Persistent SGD in the two-cluster model, with
  $\alpha=2$, $\Delta=0.5$, $\lambda=0$, $1/\tau=0.6$ and different batch sizes
  $b=1,0.3,0.1$. The continuous lines mark the numerical solution
  of DMFT equations, while the symbols are the results of simulations
  at $d=500$, $\eta=0.2$, and $R=0.01$. The dashed grey line marks the
  Bayes-optimal error from \cite{mignacco2020role}. \textbf{Right:} Generalization error as a function of the training time for full-batch gradient descent in the two-cluster model with different regularization $\lambda=0,0.1,1$ and the same parameters as in the left panel. In each panel, the inset shows the training accuracy as a function of the training time.}\label{2clusters}
\end{figure}

The left panel of Fig.~\ref{2clusters} shows the learning dynamics of
the Persistent-SGD algorithm in the two-cluster model without regularization $\l=0$. We
clearly see a good match between the numerical simulations and the
theoretical curves obtained from DMFT, notably also for small values
of batchsize $b$ and dimension $d=500$. The figure shows
that there exist regions in control parameter space where Persistent SGD is able
to reach 100\% training accuracy, while the generalization error is
bounded away from zero. Remarkably, we observe that the additional
noise introduced by decreasing the batch size $b$ results in a shift
of the early-stopping minimum of the generalization error at larger
times and that, on the time window we show, a batch size smaller than one has
a beneficial effect on the generalization error at long times. The
right panel illustrates the role of regularization in the same model
trained with full-batch gradient descent, presenting that
regularization has a similar
influence on the learning curve as small batch-size but without the
slow-down incurred by Persistent SGD. 


\begin{figure}[t]
\includegraphics[scale=0.2]{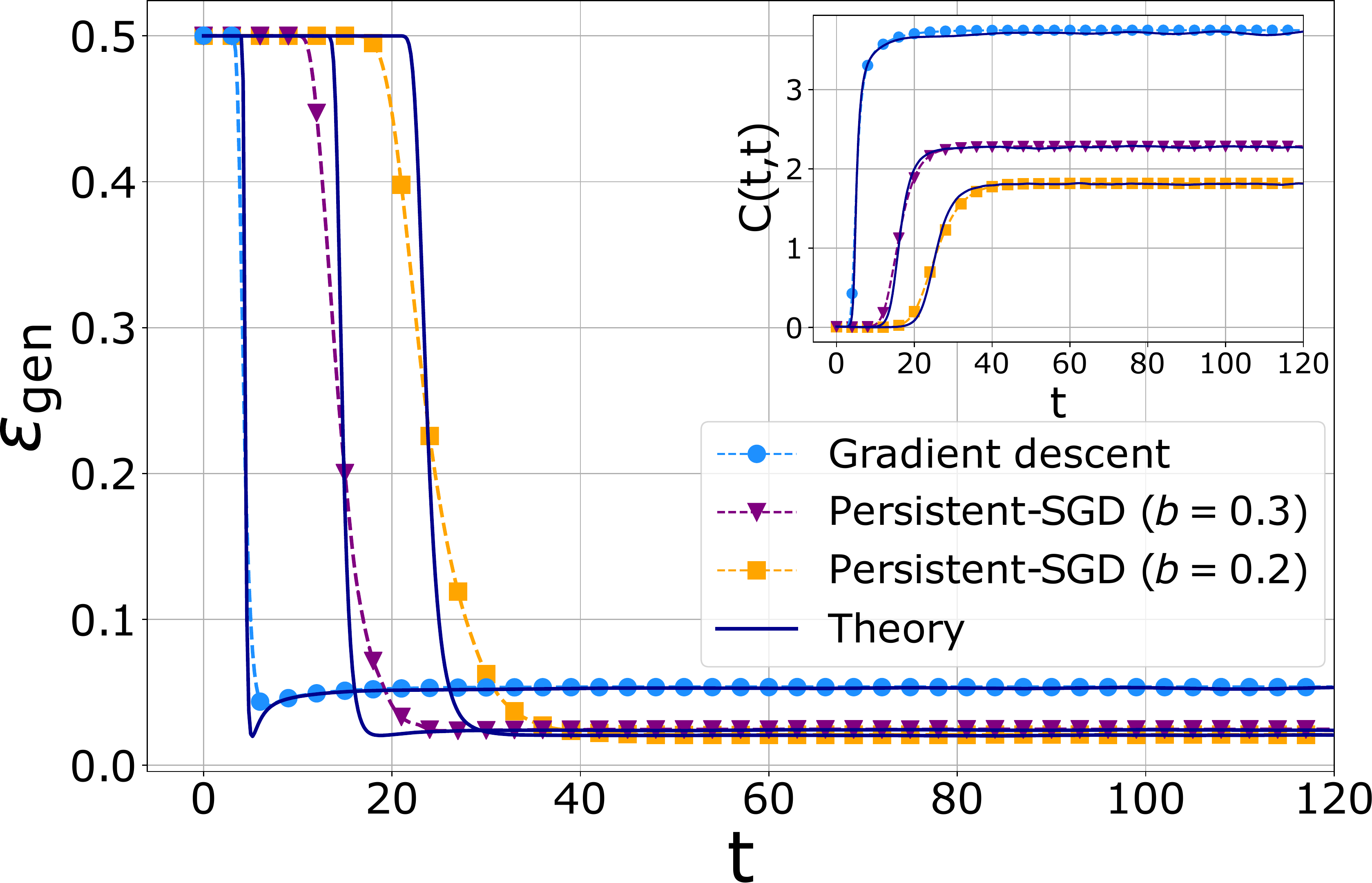} 
\includegraphics[scale=0.2]{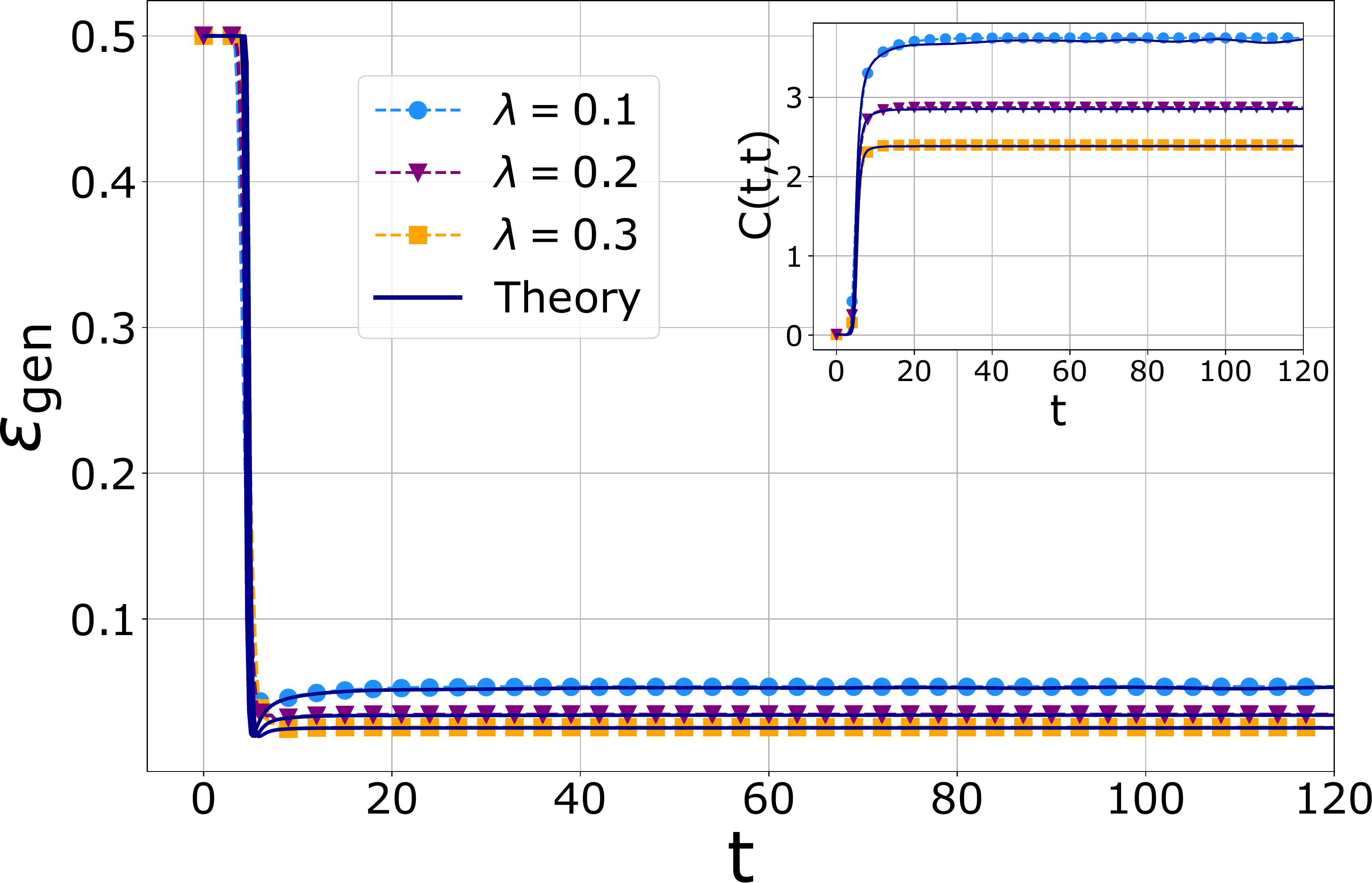} 
\caption{ \textbf{Left:} Generalization error as a function of the
  training time in the three-cluster model, at fixed $\alpha=3$,
  $\Delta=0.05$, $L=0.7$, $\lambda=0.1$, for full-batch gradient
  descent and Persistent SGD with different batch size $b=0.2,0.3$ and activation rate $1/\tau=b$. The continuous lines mark
  the numerical solution of DMFT equations, the symbols represent
  simulations at $\eta=0.2$, $R=0.01$, and $d=5000$.  \textbf{Right:} Generalization error as a function of training time for full-batch gradient descent in the three-cluster model, at fixed  $\alpha=3$, $\Delta=0.05$, $L=0.7$, $\eta=0.2$, $R=0.01$, and different regularization $\lambda=0.1,0.2,0.3$. The simulations are done at $d=5000$. In each panel, the inset shows the norm of the weights as a function of the training time. }\label{3clusters}
\end{figure}
The influence of the batch size $b$ and the regularization $\lambda$
for the three-cluster model is shown in Fig.~\ref{3clusters}. We see
an analogous effect as for the two-clusters in Fig.~\ref{2clusters}. 
In the inset of Fig.~\ref{3clusters}, we show the norm of the weights 
as a function of the training time. Both with the smaller
mini-batch size and larger regularization the norm is small,
testifying further that the two play a similar role in this case.

\begin{figure}[t]
\includegraphics[scale=0.2]{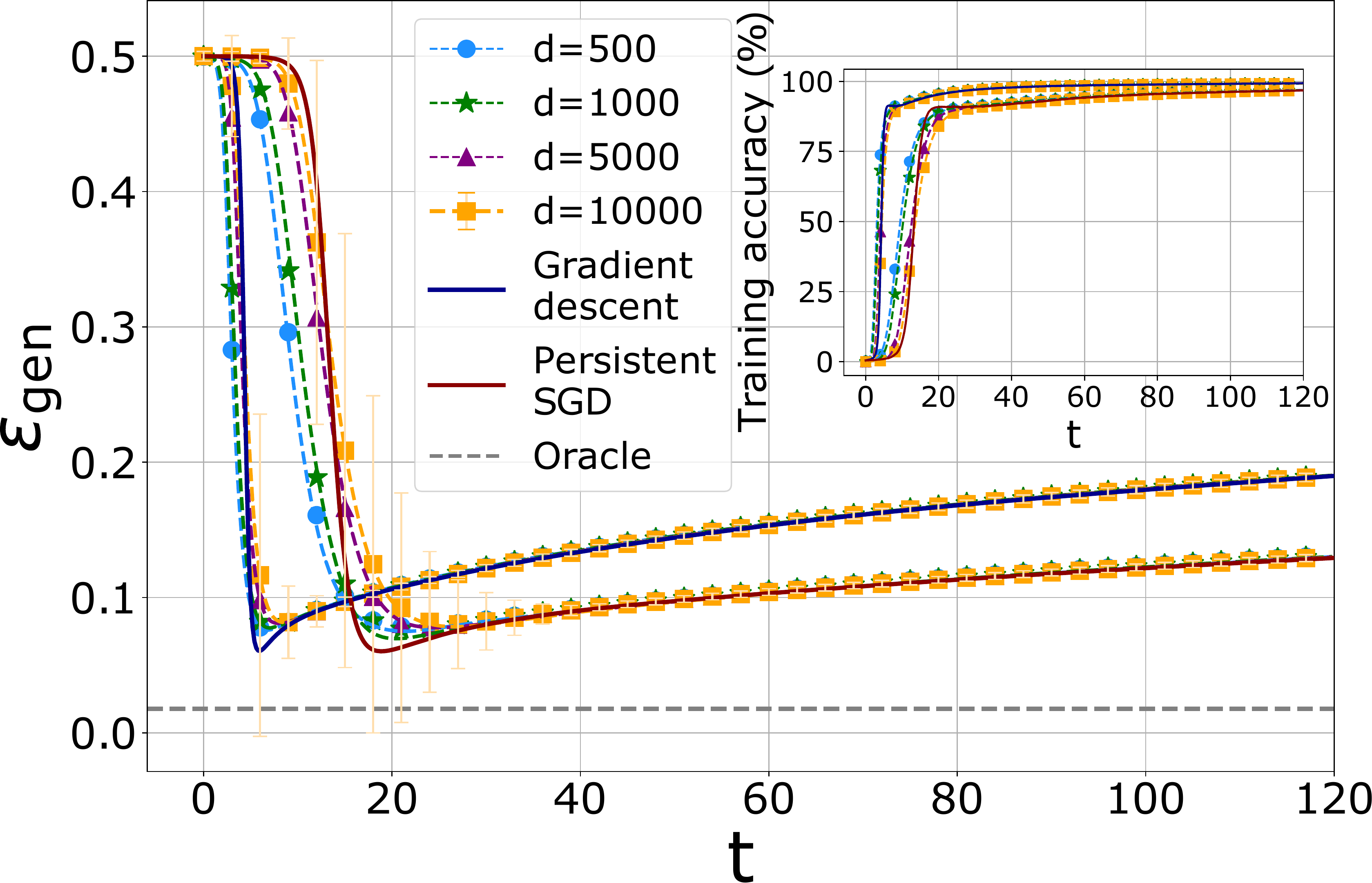} 
\includegraphics[scale=0.2]{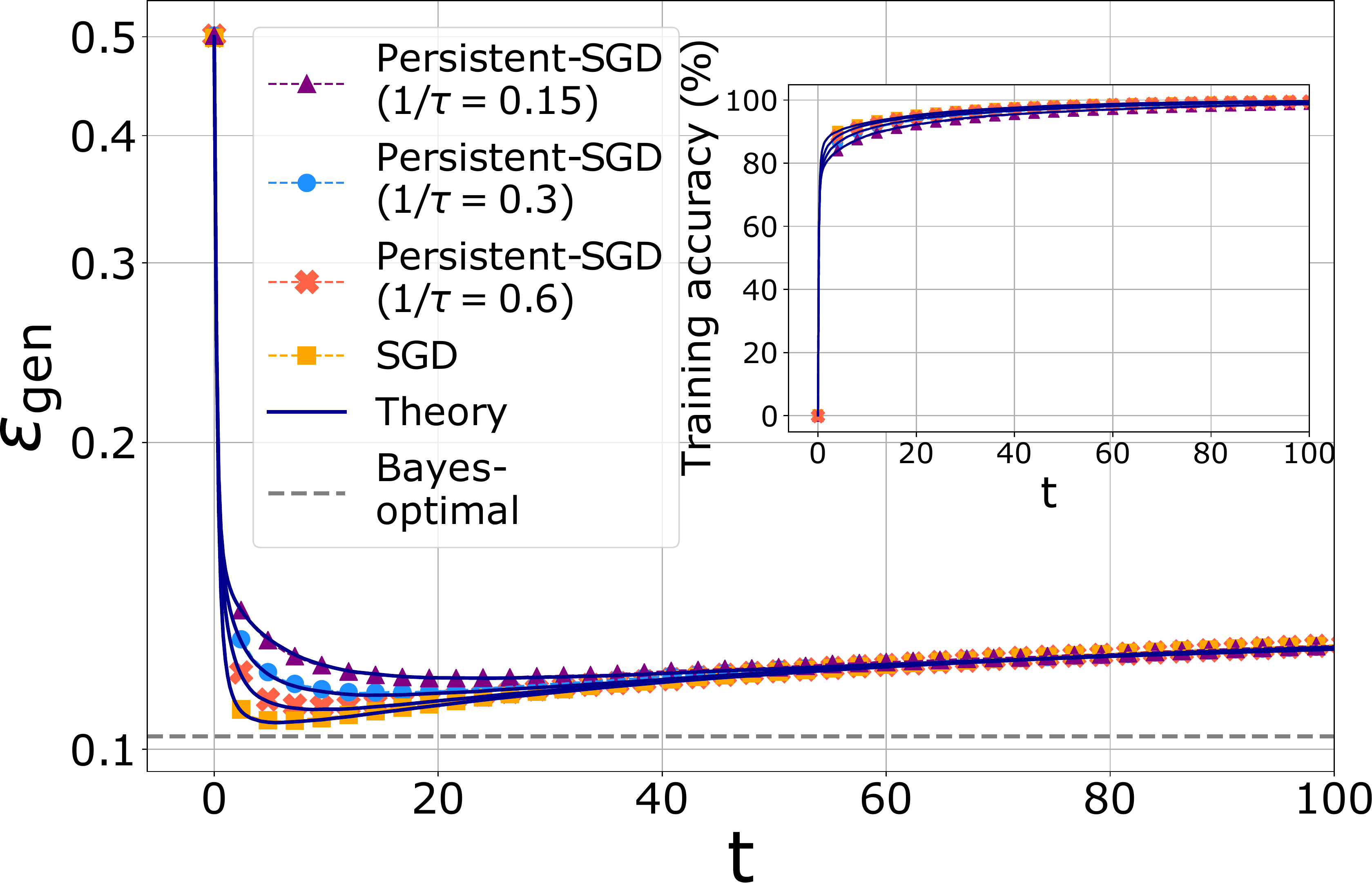} 
\caption{ \textbf{Left:} Generalization error as a function of the
  training time for full-batch gradient descent and Persistent SGD
  with $1/\tau=b=0.3$ in the three-cluster model, at fixed $\alpha=2$,
  $\Delta=0.05$, $L=0.7$ and $\lambda=0$. The continuous lines mark
  the numerical solution of DMFT equations, the symbols represent
  simulations at $\eta=0.2$, $R=1$, and increasing dimension
  $d=500,1000,5000,10000$. Error bars are plotted for $d=10000$. The dashed lines mark the oracle error (see
  supplementary material). \textbf{Right:} Generalization error as a function of
  the training time for Persistent SGD with different activation rates
  $1/\tau=0.15,0.3,0.6$ and classical SGD in the two-cluster model,
  both with $b=0.3$, $\alpha=2$, $\Delta=0.5$, $\lambda=0$,
  $\eta=0.2$, $R=0.01$. The continuous lines mark the numerical
  solution of DMFT equations (in case of SGD we use the SGD-inspired
  discretization), while the symbols represent simulations at $d=500$. The dashed lines mark the Bayes-optimal error from \cite{mignacco2020role}. In each panel, the inset displays the training accuracy as a function of time.}
\label{mixed}
\end{figure}

One difference between the two-cluster an the three-cluster models we
observe concerns the behavior of the generalization error at small
times. Actually, for the three-cluster model, good generalization is reached because of finite-size effects. Indeed, the corresponding loss function
displays a ${\mathbb{Z}_2}$ symmetry according to which for each local minimum ${\bf w}$ there is another one $-{\bf w}$ with exactly the same properties. 
Note that this symmetry is inherited from the activation function
$\phi$ \eqref{Phi_def}, which is even. This implies that if $d\to \infty$, the
generalization error would not move away from $0.5$ in finite time. However, when $d$ is large but finite, at time $t=0$ the weight vector has a finite projection on ${\bf v}^*$ which is responsible for the dynamical symmetry breaking and eventually for a low generalization error at long times. 
In order to obtain an agreement between the theory and simulations, we
initialize $m(t)$ in the DMFT equations with its corresponding
finite-$d$ average value at $t=0$. In the left panel of Fig.~\ref{mixed}, we
show that while this produces a small discrepancy at intermediate
times that diminishes with growing size, at long times the DMFT tracks perfectly the evolution of the algorithm. 

The right panel of Fig.~\ref{mixed} summarizes the effect of the
characteristic time $\tau$ in the Persistent SGD, related to the typical
persistence time of each pattern in the training mini-batch. When
$\tau$ decreases, the Persistent SGD algorithm is observed to be getting a better
early-stopping generalization error and the dynamics gets closer to
the usual SGD dynamics. As expected, the $\tau\to \eta/b$ limit
of the Persistent SGD converges to the SGD.  The SGD-inspired discretization of the DMTF equations shows a perfect agreement with the numerics.


\begin{figure}[t]
\includegraphics[scale=0.2]{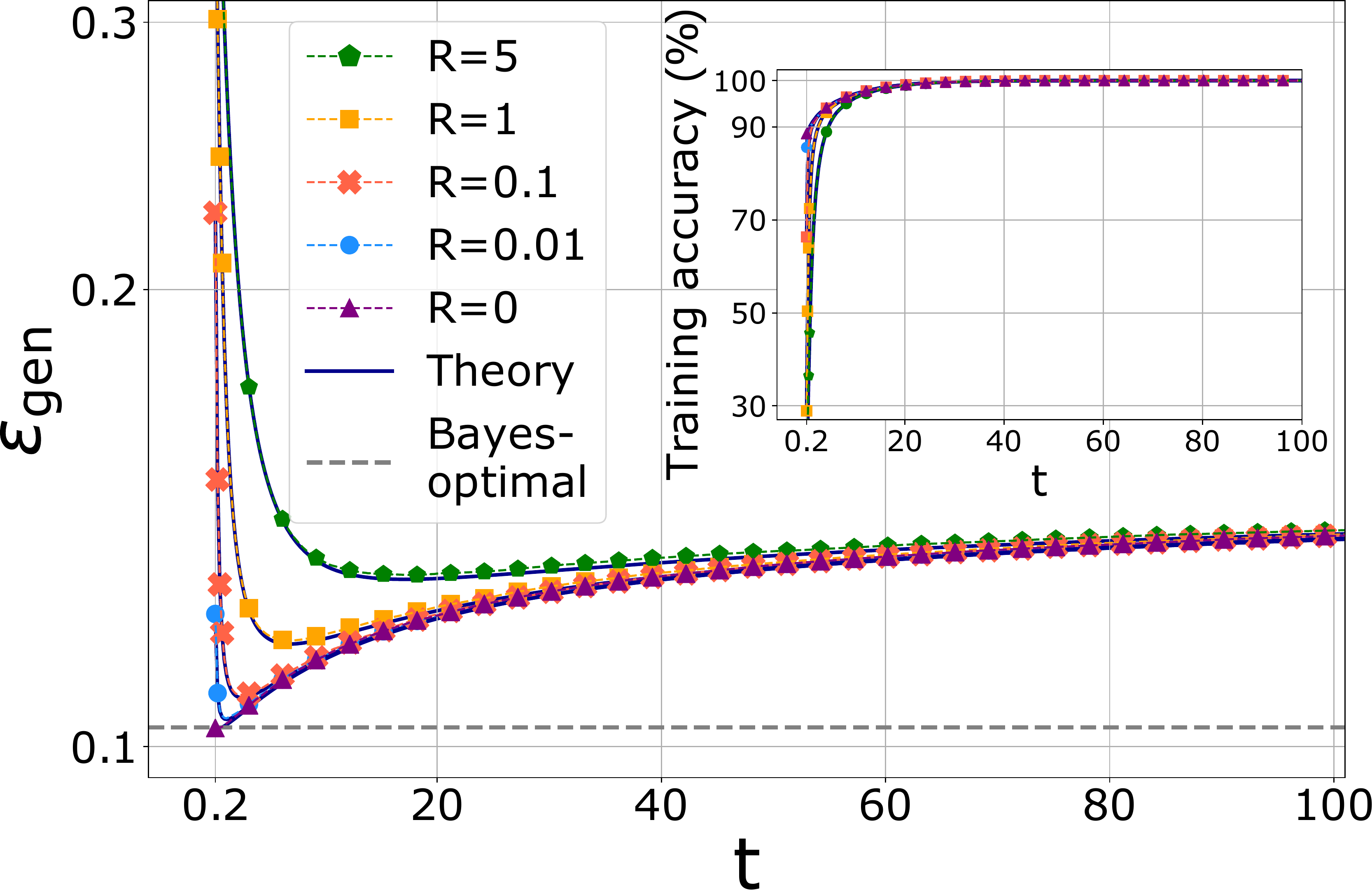} 
\includegraphics[scale=0.2]{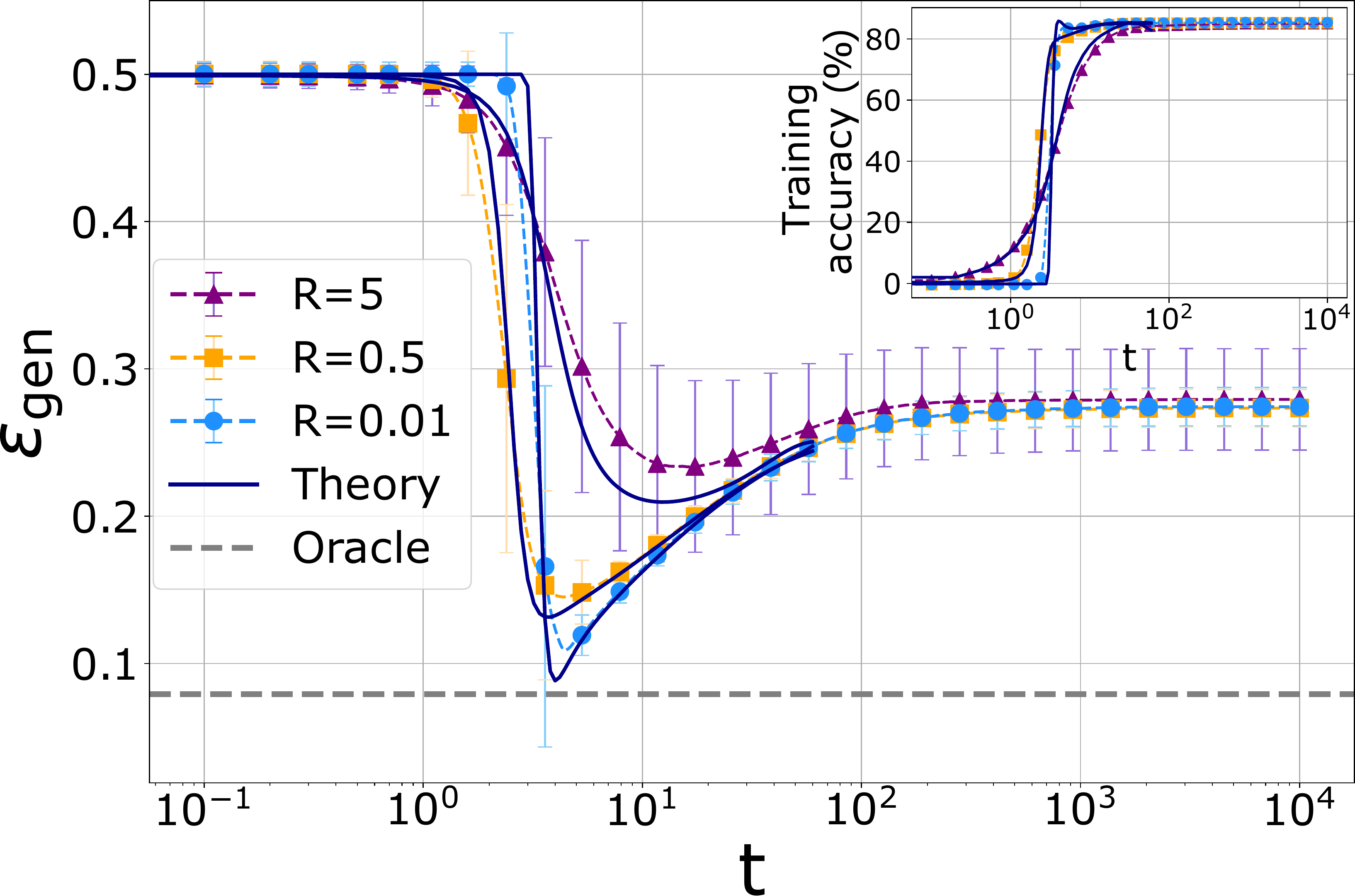} 
\caption{\textbf{Left:} Generalization error as a function of training time for full-batch gradient descent in the two-cluster model, at fixed $\alpha=2$, $\Delta=0.5$, $\lambda=0$, $\eta=0.2$, and different initialization variances $R=0,0.01,0.1,1,5$. The continuous lines mark the numerical solution of DMFT equations, while the symbols represent simulations at $d=500$. The dashed lines mark the Bayes-optimal error from \cite{mignacco2020role}. The $y-$axis is cut for better visibility. \textbf{Right:} Generalization error as a function of training time for full-batch gradient descent in the three-cluster model, at fixed $\alpha=3$, $\Delta=0.1$, $\lambda=0$, $\eta=0.1$ and different initialization variances $R=0.01, 0.5, 5$. The continuous lines mark the numerical solution of DMFT equations, while the symbols represent simulations at $d=1000$. The dashed grey line marks the oracle error (see supplementary material). In each panel, the inset shows the training accuracy as a function of time.}
\label{Role_of_R}
\end{figure}

Fig.~\ref{Role_of_R} presents the influence of the weight norm at
initialization $R$ on the dynamics, for the two-cluster (left) and
three-cluster (right) model. 
For the two-cluster case, the gradient descent algorithm with
all-zeros initialization ``jumps'' on the Bayes-optimal error at the
first iteration as derived in \cite{mignacco2020role}, and in this particular setting
the generalization error is monotonically increasing in time. As $R$
increases the early stopping error gets worse. At large times all the
initializations converge to the same value of the error, as they must,
since this is a full-batch gradient descent without regularization
that at large times converges to the max-margin estimator according
to \cite{rosset2004margin}. For the three-cluster model we observe a qualitatively similar behavior.

\newpage


\begin{ack}
This work was supported by ''Investissements d'Avenir'' LabExPALM
(ANR-10-LABX-0039-PALM), the ERC under the European Union’s Horizon
2020 Research and Innovation Program 714608-SMiLe, as well as by the French Agence Nationale de la Recherche under grant ANR-17-CE23-0023-01
PAIL and ANR-19-P3IA-0001 PRAIRIE.
\end{ack}
\newpage
\appendix

\numberwithin{equation}{section}
\section{Derivation of the dynamical mean-field equations}
The derivation of the self-consistent stochastic process discussed in the main text can be obtained using tools 
of statistical physics of disordered systems. 
In particular, it has been done very recently for a related model, the
spherical perceptron with random labels, in \cite{ABUZ18}.
Our derivation extends the known DMFT equations by including
\begin{itemize}
\item structure in the data;
\item a stochastic version of gradient descent as discussed in the main text;
\item the relaxation of the spherical constraint over the weights and the introduction of a Ridge regularization term.
\end{itemize}
There are at least two ways to write the DMFT equations. One is by using field-theoretical techniques; otherwise one can employ a dynamical version
of the so-called \emph{cavity method} \cite{MPV87}.
Here we opt for the first option that is generically very compact and immediate and it has a form that resembles very much a \emph{static} treatment of the Gibbs measure of the problem \cite{Ku02}.
We use a supersymmetric (SUSY) representation to derive the  dynamical mean-field (DMFT) equations \cite{ABUZ18,Ku92}.
We do not report all the details, that can be found in \cite{ABUZ18} along with an alternative derivation based on the cavity method, but we limit ourselves to provide the main points.
We first consider the dynamical partition function, corresponding to Eq. \eqref{Zdyn} in the main text
\begin{equation}
\begin{split}
Z_{\rm dyn}&= \left\langle\int \left[ \frac{\de {\bf w}^{(0)}}{({2\pi})^{\frac d2}}e^{-\frac{1}{2}\Vert{\bf w}^{(0)}\Vert_2^2}\right]\int_{{\bf w}(0)={\bf w}^{(0)}} \mathcal{D}{\bf w}(t)\right.\\
&\left.\times \prod_{j=1}^d \,\delta\left[-\dot {\rm w}_j(t)-\lambda {\rm w}_j(t) -\sum_{\mu=1}^n \,s_{\mu}(t) \L'\left( y_{\mu}, \frac{{\bf w}(t)^\top {\bf x}_{\mu}}{\sqrt{d}}\right)\frac{{\rm x}_{\mu,j}}{\sqrt{d}}\right] \right\rangle,
\end{split}\label{Zdyn_supmat}
\end{equation}
where the brackets $\langle\cdot\rangle$ stand for the average over $s_{\mu}(t)$, $ y_\mu$ and the realization of the noise in the training set. The average over the initial condition is written explicitly. Note that we choose an initial condition that is Gaussian, but we could have chosen a different probability measure over the initial configuration of the weights. The equations can be generalized to other initial conditions as soon as they do not depend on quenched random variables that enter in the stochastic gradient descent (SGD) dynamics and their distribution is separable. As observed in the main text, we have that $Z_{\rm dyn}=\langle Z_{\rm dyn} \rangle=1$. We can write the integral representation of the Dirac delta function in Eq. \ref{Zdyn_supmat} by introducing a set of fields $\hat{\bfw}(t)$
\begin{equation}
\begin{split}
Z_{\rm dyn}= \left\langle\int \mathcal{D}{\bf w}(t)\mathcal{D}\hat {\bf w}(t) \,e^{S_{\rm dyn}}\right\rangle,
\end{split}
\end{equation} 
where the dynamical action $S_{\rm dyn}$ is defined as in Eq. \eqref{Sdyn} of the main text
\begin{equation}
S_{\rm dyn}=\sum_{j=1}^d \int_0^{+\infty}\dd t \,i{\hat{\rm w}_j(t)}\left( -\dot{\rm w}_j(t)-\lambda {\rm w}_j(t) -\sum_{\mu=1}^n \,s_{\mu}(t) \L'\left( y_{\mu}, \frac{{\bf w}(t)^\top {\bf x}_{\mu}}{\sqrt{d}}\right)\frac{{\rm x}_{\mu,j}}{\sqrt{d}}\right).\label{Sdyn_supmat}
\end{equation}
\subsection{SUSY formulation}
The dynamical action $S_{\rm dyn}$ \eqref{Sdyn_supmat} can be rewritten in a supersymmetric form, by extending the time coordinate to include two Grassman coordinates $\theta$ and $\bar{\theta}$, i.e. $t_a \rightarrow a = (t_a,\theta_a,\bar{\theta}_a)$. The dynamic variable ${\bf w}(t_a)$ and the auxiliary variable $i{\bf \hat{w}}(t_a)$ are encoded in a super-field
\begin{equation}
{\bf w}(a)={\bf w}(t_a)+i\,\theta_a \bar{\theta}_a  {\bf \hat{w}}(t_a).
\end{equation}
From the properties of Grassman variables \cite{Zi96}
\begin{equation}
\begin{split}
\theta^2=\bar{\theta}^2=\theta\bar{\theta}+\bar{\theta}\theta=0 , \\
\int \dd \theta = \int \dd \bar{\theta}=0 ,\qquad \int \dd \theta \,\theta= \int \dd \bar{\theta} \,\bar{\theta}=1 , \\
\partial_{\theta}g(\theta)=\int \dd \theta \,g(\theta) \quad \text{ for a generic function } g,
\end{split}
\end{equation}
it follows that
\begin{equation}
\int \dd a \,f \left({\bf w}(a)\right) = \int_0^{+\infty} \dd t_a\, i {\bf \hat{w}}(t_a) f'\left({\bf w}(t_a)\right).\label{Grass_f_supmat}
\end{equation}
We can use Eq. (\ref{Grass_f_supmat}) to rewrite $S_{\rm dyn}$. We obtain
\begin{equation}
\begin{split}
S_{\rm dyn}=-\frac{1}{2} \int \dd a \dd b\, \mathcal{K}(a,b) {\bf w}(a)^\top {\bf w}(b)-\sum_{\mu=1}^n\int \dd a\,s_{\mu}(a)\, \L\left( y_\mu,h_{\mu}(a) \right)
,\end{split}
\end{equation}
where we have defined $h_{\mu}(a)\equiv {\bf w}(a)^\top {\bf x}_{\mu}/\sqrt{d}$ and we have implicitly defined the kernel $\KK(a,b)$ such that
\beq
-\frac{1}{2} \int \dd a \dd b\, \mathcal{K}(a,b) {\bf w}(a)^\top {\bf w}(b) = \sum_{j=1}^d \int_0^{+\infty}\dd t \,i{\hat{\rm w}_j(t)}\left( -\dot{\rm w}_j(t)-\lambda {\rm w}_j(t)\right)\:.
\eeq
By inserting the definition of $h_{\mu}(a)$ in the partition function, we have
\begin{equation}
\begin{split}
Z_{\rm dyn}&=\left\langle\int \mathcal{D}{\bf w}(a)
  \mathcal{D}h_{\mu}(a) \mathcal{D}\hat{h}_{\mu}(a)
  \,\exp\left[-\frac{1}{2} \int \dd a \dd b\, \mathcal{K}(a,b) {\bf
      w}(a)^\top {\bf w}(b)\right. \right.\\
&\left.\left.-\sum_{\mu=1}^n\int \dd a\,s_{\mu}(a)\, \L \left(y_\mu, h_{\mu}(a) \right)\right] \exp\left[\sum_{\mu=1}^n \int \dd a \,i\,\hat{h}_{\mu}(a)\left(h_{\mu}(a)- \frac{{\bf w}(a)^\top {\bf x}_{\mu}}{\sqrt{d}}\right)\right]\right\rangle.
\end{split} 
\label{Zdyn_a_supmat}
\end{equation}
Let us consider the last factor in the integral in (\ref{Zdyn_a_supmat}). We can perform the average over the random vectors ${\bf z}_\mu\sim \mathcal{N}({\bf 0}, {\bf I}_d)$, denoted by an overline, as
\begin{equation}
\begin{split}
&\overline{\exp\left[\sum_{\mu=1}^n\int \dd a \, i\,\hat{h}_{\mu}(a)\left(h_{\mu}(a)-\frac{{\bf w}(a)^\top {\bf x}_{\mu}}{\sqrt{d}}\right)\right]}\\
&=\overline{\exp\left[\sum_{\mu=1}^n \int \dd a \,i\,\hat{h}_{\mu}(a)\left(h_{\mu}(a)-c_\mu m(a) - \sqrt{\frac{\Delta}{d}}{\bf w}(a)^\top {\bf z}_{\mu}\right)\right]}\\
&=\exp \left[\sum_{\mu=1}^n \int \dd a \,i\,\hat{h}_{\mu}(a)\left(h_{\mu}(a)-c_\mu m(a) \right)-\frac{\Delta}{2}\sum_{\mu=1}^n \int \dd a\,\dd b \, Q(a,b) \hat{h}_{\mu}(a)\hat{h}_{\mu}(b)\right],
\end{split}
\end{equation}

where we have defined
\begin{equation}
\begin{split}
m(a)&=\frac{1}{d}{\bf w}(a)^\top {\bf v^\ast},\\
Q(a,b)&=\frac{1}{d}{\bf w}(a)^\top{\bf w}(b).
\end{split}
\end{equation}

By inserting the definitions of $m(a)$ and $Q(a,b)$ in the partition function, we obtain
\begin{equation}
Z_{\rm dyn} = \int \mathcal{D}{\bf Q}\,\mathcal{D}{\bf m}\,\, e^{d S({\bf Q},{\bf m})},
\label{large_D_supmat}
\end{equation}
where ${\bf Q}=\{Q(a,b)\}_{a,b}\, ,$ ${\bf m}=\{m(a)\}_a$ and
\begin{equation}
\begin{split}
S({\bf Q},{\bf m})&=\frac{1}{2}\log \det \left(Q(a,b)-m(a)m(b)\right)-\frac{1}{2}\int \dd a \dd b \, \mathcal{K}(a,b) Q(a,b) +\alpha \log \mathcal{Z}, \\
\mathcal{Z}&=\left\langle\int \mathcal{D}h(a)\mathcal{D}\hat{h}(a)
  \,\exp\left[-\frac{\Delta}{2}\int \dd a \dd b \,\,Q(a,b)
    \hat{h}(a)\hat{h}(b) \right.\right.\\
&\left.\left.+ \int \dd a \, i \hat{h}(a)\left(h(a)-c m(a)\right)-\int \dd a \,s(a)\,\L \left(y, h(a)\right)\right]\right\rangle\:.
\end{split}
\end{equation}
We have used that the samples are i.i.d. and removed the index $\mu=1,...n$.
The brackets denote the average over the random variable $c$, that has the same distribution as the $c_\mu$, over $y$, distributed as $y_{\mu}$, and over the random process of $s(t)$, defined by Eq. \eqref{PSGD} in the main text. 
If we perform the change of variable $Q(a,b)\leftarrow Q(a,b) +m(a)m(b)$, we obtain
\begin{equation}
\begin{split}
S({\bf Q},{\bf m})&=\frac{1}{2}\log \det Q(a,b)-\frac{1}{2}\int \dd a \dd b \, \mathcal{K}(a,b) \left(Q(a,b)+m(a)m(b)\right) +\alpha \log \mathcal{Z}, 
\\
\mathcal{Z}&=\left\langle \int \mathcal{D}h(a)\mathcal{D}\hat{h}(a) \,e^{S_{\rm loc}}\right\rangle,
\end{split}
\end{equation}
where the effective local action $S_{\rm loc}$ is given by
\beq
\begin{split}
S_{\rm loc} &=-\frac{\Delta}{2}\int \dd a \dd b \,\,Q(a,b)\hat{h}(a)\hat{h}(b) -\frac{\Delta}{2}\left(\int \dd a \, \hat{h}(a)m(a)\right)^2\\
&+ \int \dd a \, i \hat{h}(a)\left(h(a)-c m(a)\right)-\int \dd a \,s(a)\,\L \left( y,h(a)\right).
\end{split}
\eeq
Performing a Hubbard-Stratonovich transformation on $\exp \left[ -\frac{\Delta}{2}\left(\int \dd a \, \hat{h}(a)  m(a)\right)^2\right]$ and a set of transformations on the fields $h(a)$, we obtain that we can rewrite $\ZZ$ as
\beq
\begin{split}
\mathcal{Z}&=\left\langle\int \frac{\dd h_0}{\sqrt{2\pi}}e^{-\frac{h_0^2}{2}}\int \mathcal{D}h(a)\mathcal{D}\hat{h}(a) \,\exp\left[-\frac{1}{2}\int \dd a \dd b \,\,Q(a,b)\hat{h}(a)\hat{h}(b) \right.\right.\\ 
&\left.\left.+ \int \dd a \, i \hat{h}(a)h(a)-\int \dd a \,s(a)\,\L \left( y, \sqrt\D h(a)+m(a)(c+\sqrt{\Delta}h_0 )\right)\right]\right\rangle.
\end{split}
\eeq
\subsection{Saddle-point equations}
We are interested in the large $d$ limit of $Z_{\rm dyn}$, in which, according to Eq.~\eqref{large_D_supmat}, the partition function is dominated by the saddle-point value of $S({\bf Q}, {\bf m})$:
\begin{equation}
\begin{cases}
\displaystyle
\frac{\delta S({\bf Q}, {\bf m})}{\delta Q(a,b)}\biggr\rvert _{({\bf Q},{\bf m})=({\bf \tilde{Q}},{\bf \tilde{m}})}=0\\\\
\displaystyle
\frac{\delta S({\bf Q}, {\bf m})}{\delta m(a)}\biggr\rvert _{({\bf Q},{\bf m})=({\bf \tilde{Q}},{\bf \tilde{m}})}=0
\end{cases}.\label{saddle_point_eqs}
\end{equation}
$\tilde Q(a,b)$ is obtained from the equation
\begin{equation}
-\mathcal{K}(a,b)+Q^{-1}(a,b)+\frac{2\alpha}{\mathcal{Z}}\frac{\delta \mathcal{Z}}{\delta Q(a,b)}=0.
\label{speQ}
\end{equation}
The saddle-point equation for $\tilde m(a)$ is instead
\begin{equation}
-\int \dd b \,\mathcal{K}(a,b) m(b) + \frac{\alpha}{\mathcal{Z}}\frac{\delta\mathcal{Z}}{\delta m(a)}=0.\label{speM}
\end{equation}
It can be easily shown by exploiting the Grassmann structure of Eqs. \eqref{speQ}-\eqref{speM} that they lead to a self consistent stochastic process described by 
\beq
\dot{h}(t)=-\tilde \lambda(t) h(t)- \sqrt \D s(t)\L'\left(y,r(t) - Y(t)\right)+\int _{0}^t \dd t' M_R(t,t')h(t')+\xi(t),\label{eff_process_supmat}
\eeq
where the initial condition is drawn from $P(h(0)) \sim e^{- h(0)^2/(2R)}/\sqrt{2\pi}$, and $r(t)=\sqrt \D h(t)+ m(t)(c+\sqrt{\Delta}h_0 )$, with $P_0(h_0) \sim e^{-h_0^2/2}/\sqrt{2\pi}$. We have defined the auxiliary functions
\beq
\begin{split}
\mu(t)&=\alpha \left< s(t) \left(c+\sqrt \D h_0\right)\L' \left(y, r(t)\right)\right>,\\
\hat{\lambda}(t)&=\alpha \Delta\left<s(t)\L'' \left(y, r(t)\right)\right>,\\
\tilde \lambda(t) & =\lambda+\,\hat{\lambda}(t),\\
\end{split}\label{auxiliary_supmat}
\eeq
and kernels
\beq
\begin{split}
M_C(t,t')&=\alpha\D\left< s(t)s(t')\L' \left(y,r(t)\right)\L' \left(y,r(t')\right)\right>,
\\
M_R(t,t')&=\alpha \D^{3/2} \left<s(t) s(t')\L' \left(y, r(t)\right) \L'' \left(y, r(t')\right)\, i\hat{h}(t')\right> \\
&\equiv \left.\a  \D \frac{\delta }{\delta Y(t')} \langle s(t) \L'(y,r(t))\rangle\right|_{Y=0}.\label{kernels_supmat}
\end{split}
\eeq
In addition, from \eqref{speM} , one can derive an ordinary differential equation for the magnetization
\beq
\dot m (t) = -\lambda {m}(t)-\mu(t).\label{ode_m_supmat}
\eeq

The brackets in the previous equations denote, at the same time, the average over the label $y$, the process $s(t)$, as well as the average over the noise $\xi(t)$ and both $h_0$ and $h(0)$, whose probability distributions are given by $P(h(0))$ and $P_0(h_0)$ respectively.
In other words, one has a set of kernels, such as $M_R(t,t')$ and $M_C(t,t')$, that
can be obtained as an average over the stochastic process for $h(t)$ and therefore must be computed self-consistently.

Finally, Eq.~\eqref{speQ} gives rise to Eq.~(20) of the main text while Eq.~\eqref{speM} gives rise to the equation for the evolution of the magnetization. 
Note that the norm of the weight vector $\bfw(t)$ can be also computed by sampling the stochastic process
\beq
\begin{split}
\dot {\rm w}(t) &= -\tilde \lambda(t) {\rm w}(t) +\int_0^t \de t' M_R(t,t')({\rm w}(t')-m(t') h_0) +\xi(t)+ h_0(\hat{\l}(t) m(t)-\mu(t)),\\
P({\rm w}_0)&=\frac{1}{\sqrt{2\pi R}}e^{-{\rm w}_0^2/(2R)},
\end{split}
\eeq
from which one gets
\beq
C(t,t') =\langle {\rm w}(t)^2\rangle\:.
\eeq

\subsection{Numerical solution of DMFT equations}
The algorithm to solve the DMFT equations that are summed up in Eq.~\eqref{eff_process_supmat} is the most natural one.
It can be understood in the following way. The outcome of the DMFT is the computation of the kernels and functions appearing in it, namely
$m(t)$, $M_C(t,t')$ and so on.
They are determined as averages over the stochastic process that is defined through them.
Therefore, one needs to solve the system of equations in a self-consistent way. 
The straightforward way to do that is to proceed by iterations:
\begin{enumerate}
\item We start from a random guess of the kernels, that we use to sample the stochastic process \eqref{eff_process_supmat} several times;
\item We compute the averages over these multiple realizations to obtain the updates of the auxiliary functions \eqref{auxiliary_supmat} and kernels \eqref{kernels_supmat}, along with the magnetization \eqref{ode_m_supmat};
\item We use these new guesses to sample again multiple realizations of the stochastic process;
\item We repeat steps 2. and 3. until the kernels reach a fixed point.
\end{enumerate} 
As in all iterative solutions of fixed point equations, it is natural to introduce some damping in the update of the kernels to avoid wild oscillations. Note that the DMFT fixed point equations are deterministic, hence at given initial condition the solution is unique. Indeed, the kernels computed by DMFT are causal and a simple integration scheme of the equations is just extending them progressively in time starting from their initial value, which is completely deterministic given the initial condition for the stochastic process.
This procedure has been first implemented in \cite{EO92,EO94} and recently developed further in other applications \cite{RBBC19, MSZ20}.
However, DMFT has a long tradition in condensed matter physics \cite{GKKR96} where more involved algorithms have been developed. 

\section{Generalization error}
The generalization error at any time step is defined as the fraction of mislabeled instances:
\begin{equation}
\varepsilon_{\rm gen}(t)\equiv\frac{1}{4}\mathbb{E}_{{\bf X},{\bf y},{\bf x}_{\rm new},y_{\rm new}}\left[\left(y_{\rm new}-\hat{y}_{\rm new}\left({\bf w}(t)\right)\right)^2\right],\label{egen_supmat}
\end{equation}
where $\{{\bf X},{\bf y}\}$ is the training set, ${\bf x}_{\rm new}$ is an unseen data point and $\hat{y}_{\rm new}$ is the estimator for the new label $y_{\rm new}$. The dependence on the training set here is hidden in the weight vector ${\bf w}(t)={\bf w}(t,{\bf X}, {\bf y})$.
\subsection{Perceptron with linear activation function}\label{section:gen_sign_supmat}
In this case, the estimator for a new label is $\hat{y}_{\rm new}\left({\bf w}(t)\right)=\sign\left({\bf w}(t)^\top {\bf x}_{\rm new}\right)$. 
The generalization error in the infinite dimensional limit $d\rightarrow\infty$ has been computed in \cite{mignacco2020role} and reads
\begin{equation}
\varepsilon_{\rm gen}(t)=\frac{1}{2}\erfc\left(\frac{m(t)}{\sqrt{2\Delta \, C(t,t)}}\right).\label{egen_sign_supmat}
\end{equation}

\subsection{Perceptron with door activation function}
In this case, the estimator for a new label is $\hat{y}_{\rm new}\left({\bf w}(t)\right)=\sign\left(\frac{1}{d}({\bf w}(t)^\top {\bf x}_{\rm new})^2-L^2\right)$. From Eq. \eqref{egen_supmat}, we have that
\begin{equation}
\varepsilon_{\rm gen}(t)=\frac{1}{2}\left(1-\mathbb{E}_{{\bf X},{\bf y},{\bf x}_{\rm new},  y_{\rm new}}\left[ y_{\rm new}\cdot\hat{y}_{\rm new}({\bf w}(t))\right]\right).\label{egen_split_supmat}
\end{equation}
We consider the second term of \eqref{egen_split_supmat}
\begin{equation}
\begin{split}
\mathbb{E}_{{\bf X},{\bf y},{\bf x}_{\rm new}, y_{\rm new}}\left[ y_{\rm new}\cdot\hat{y}_{\rm new}({\bf w}(t))\right]=\mathbb{E}_{{\bf X},{\bf y},{\bf x}_{\rm new}}\left[\sign\left(\frac{y_{\new}}{d}({\bf w}(t)^\top {\bf x}_{\rm new})^2-y_{\new}L^2\right)\right].
\end{split}
\end{equation}
In the high dimensional limit, the overlap between weight vector and data point at each time step concentrates 
\begin{equation}
\begin{split}
\frac{{\bf w}(t)^\top {\bf x}_{\new}}{\sqrt{d}}=\frac{{\bf w}(t)^\top}{\sqrt{d}}\left(c_{\new}\frac{{\bf v^*}}{\sqrt{d}}+\sqrt{\Delta}\,{\bf z}_{\new}\right)
\underset{d\rightarrow\infty}{\rightarrow}c_{\rm new}\,m(t) + \sqrt{\Delta C(t,t)}\,z,
\end{split}
\end{equation}
where $z\sim \mathcal{N}(0,1)$. Therefore, we obtain
\begin{equation}
\begin{split}
\mathbb{E}_{{\bf X},{\bf y},{\bf x}_{\rm new}, y_{\rm new}}\left[ y_{\rm new}\cdot\hat{y}_{\rm new}({\bf w}(t))\right]\simeq\\\simeq\mathbb{E}_{c_{\rm new}, z,y_{\rm new}}\left[\sign\left(y_{\rm new}\left(c_{\rm new}\,m(t) + \sqrt{\Delta C(t,t)}\,z\right)^2-y_{\new}L^2\right)\right]\\
=\mathbb{P}\left(y_{\rm new}\left(c_{\rm new}\,m(t) + \sqrt{\Delta C(t,t)}\,z\right)^2\geq y_{\new}L^2\right)\\
-\mathbb{P}\left(y_{\rm new}\left(c_{\rm new}\,m(t) + \sqrt{\Delta C(t,t)}\,z\right)^2< y_{\new}L^2\right)
\end{split}
\end{equation}
and  the generalization error in the infinite dimensional limit $d\rightarrow\infty$ is 
\begin{equation}
\varepsilon_{\rm gen}(t)=(1-\rho) \erfc\left(\frac{L}{\sqrt{2\Delta C(t,t)}}\right)+\frac{\rho}{2}\left(\erf\left(\frac{L-m(t)}{\sqrt{2\Delta C(t,t)}}\right)+\erf\left(\frac{L+m(t)}{\sqrt{2\Delta C(t,t)}}\right)\right).
\end{equation}

\section{Oracle error}
We call \emph{oracle error} the classification error made by an ideal oracle that has access to the vector $\bf v^*$ that characterizes the centers of the clusters in the two models under consideration (see Sec. \ref{sec:two_cluster} in the main text). We define the oracle's estimator $\hat{y}_{\new}^O$ given a new data point ${\bf x}_{\new}$ as
\begin{equation}
\hat{y}_{\new}^{O}=\arg\max_{\tilde{y}_{\new}} {\rm p}\left(\tilde{y}_{\new}|{\bf x}_{\new}\right),\label{def_Oest_supmat}
\end{equation}
where the prior over the label $\tilde y_{\new}$ and the coefficient $\tilde c_{\new}$ along with the channel distribution 
\begin{equation}
{\rm p}\left({\bf x}_{\new}|\tilde{c}_{\new}\right)\propto \exp\left[-\frac{1}{2\Delta}\Vert{\bf x}_{\new}-\frac{\tilde{c}_{\new}}{\sqrt{d}}{\bf v^*}\Vert^2_2\right]
\end{equation}
are known. We can rewrite the probability in Eq. (\ref{def_Oest_supmat}) as
\begin{equation}
\begin{split}
{\rm p}\left(\tilde{y}_{\new}|{\bf x}_{\new}\right)\propto \sum_{\tilde{c}_{\new}=0,\pm1}{\rm p}\left(\tilde{y}_{\new},\tilde{c}_{\new}\right){\rm p}\left({\bf x}_{\new}|\tilde{c}_{\new}\right)=(1-\rho)\delta(\tilde{y}_{\new}+1)e^{-\frac{1}{2\Delta}\Vert {\bf x}_{\new}\Vert^2_2}\\+\frac{\rho}{2}\delta(\tilde{y}_{\new}-1)\left(e^{-\frac{1}{2\Delta}\Vert {\bf x}_{\new}-\frac{1}{\sqrt{d}}{\bf v^*}\Vert^2_2}+e^{-\frac{1}{2\Delta}\Vert {\bf x}_{\new}+\frac{1}{\sqrt{d}}{\bf v^*}\Vert^2_2}\right)\\
=e^{-\frac{1}{2\Delta}\Vert {\bf x}_{\new}\Vert^2_2}\left[(1-\rho)\delta(\tilde{y}_{\new}+1)+\rho\delta(\tilde{y}_{\new}-1)e^{-\frac{1}{2\Delta}}\cosh\left(\frac{1}{\Delta\sqrt{d}}{\bf x}_{\new}^\top{\bf v^*}\right)\right].
\end{split}\label{pnew_supmat}
\end{equation}
The oracle error is then
\begin{equation}
\varepsilon_{\rm gen}^{O}=\mathbb{P}\left(\hat{y}_{\new}^{O}\neq y_{\new}\right)=(1-\rho)\,\mathbb{P}\left(\hat{y}_{\new}^{O}=1| y_{\new}=-1\right)+\rho\,\mathbb{P}\left(\hat{y}_{\new}^{O}=-1| y_{\new}=1\right).
\end{equation}
We can compute the two terms in the above equation separately
\begin{equation}
\begin{split}
\mathbb{P}\left(\hat{y}_{\new}^{O}=1| y_{\new}=-1\right)=\mathbb{P}\left(\rho e^{-\frac{1}{2\Delta}}\cosh\left(\frac{1}{\sqrt{\Delta d}}{\bf z}_{\new}^\top {\bf v^*}\right)>1-\rho\right)\\=\mathbb{P}\left(\rho e^{-\frac{1}{2\Delta}}\cosh\left(\frac{\zeta_{\new}}{\sqrt{\Delta}}\right)>1-\rho\right)=\erfc\left(\sqrt{\frac{\Delta}{2}}\biggr\rvert\arccosh\left(\frac{(1-\rho)}{\rho}e^{1/2\Delta}\right)\biggr\rvert\right),
\end{split}
\end{equation}
and
\begin{equation}
\begin{split}
\mathbb{P}\left(\hat{y}_{\new}^{O}=-1| y_{\new}=1\right)=\mathbb{P}\left(1-\rho>\rho e^{-\frac{1}{2\Delta}}\cosh\left(\frac{c_{\new}}{\Delta}+\frac{1}{\sqrt{\Delta d}}{\bf z}_{\new}^\top {\bf v^*}\right)\right)\\
=\mathbb{P}\left(1-\rho>\rho e^{-\frac{1}{2\Delta}}\cosh\left(\frac{c_{\new}}{\Delta}+\frac{\zeta_{\new}}{\sqrt{\Delta}}\right)\right)\\=\frac{1}{2}\left[\erf\left(\frac{\Delta\biggr\rvert\arccosh\left(\frac{(1-\rho)}{\rho}e^{1/2\Delta}\right)\biggr\rvert+1}{\sqrt{2\Delta}}\right)
+\erf\left(\frac{\Delta\biggr\rvert\arccosh\left(\frac{(1-\rho)}{\rho}e^{1/2\Delta}\right)\biggr\rvert-1}{\sqrt{2\Delta}}\right)\right],\end{split}
\end{equation}
where ${\bf z}_{\new}\sim\mathcal{N}({\bf 0},{\bf I}_d)$, $\zeta_{\new}\sim\mathcal{N}(0,1)$, and $c_{\new}=\pm1$ with probability $1/2$.\\
Finally, we obtain that the oracle error is
\begin{equation}
\begin{split}
\varepsilon_{\rm gen}^{BO}=(1-\rho)\erfc\left(\sqrt{\frac{\Delta}{2}}\biggr\rvert\arccosh\left(\frac{(1-\rho)}{\rho}e^{1/2\Delta}\right)\biggr\rvert\right)\\+\frac{\rho}{2}\left[\erf\left(\frac{\Delta\biggr\rvert\arccosh\left(\frac{(1-\rho)}{\rho}e^{1/2\Delta}\right)\biggr\rvert+1}{\sqrt{2\Delta}}\right)
+\erf\left(\frac{\Delta\biggr\rvert\arccosh\left(\frac{(1-\rho)}{\rho}e^{1/2\Delta}\right)\biggr\rvert-1}{\sqrt{2\Delta}}\right)\right].
\end{split}
\end{equation}

\bibliographystyle{unsrt}
\bibliography{HS}

\begin{thebibliography}{10}

\bibitem{safran2017spurious}
Itay Safran and Ohad Shamir.
\newblock Spurious local minima are common in two-layer relu neural networks.
\newblock {\em arXiv preprint arXiv:1712.08968}, 2017.

\bibitem{LPA19}
Shengchao Liu, Dimitris Papailiopoulos, and Dimitris Achlioptas.
\newblock Bad global minima exist and sgd can reach them.
\newblock {\em arXiv preprint arXiv:1906.02613}, 2019.

\bibitem{BO97}
Siegfried B{\"o}s and Manfred Opper.
\newblock Dynamics of training.
\newblock In {\em Advances in Neural Information Processing Systems}, pages
  141--147, 1997.

\bibitem{SMG13}
Andrew~M Saxe, James~L McClelland, and Surya Ganguli.
\newblock Exact solutions to the nonlinear dynamics of learning in deep linear
  neural networks.
\newblock {\em arXiv preprint arXiv:1312.6120}, 2013.

\bibitem{SS95Short}
David Saad and Sara~A Solla.
\newblock Exact solution for on-line learning in multilayer neural networks.
\newblock {\em Physical Review Letters}, 74(21):4337, 1995.

\bibitem{SS95}
David Saad and Sara~A Solla.
\newblock On-line learning in soft committee machines.
\newblock {\em Physical Review E}, 52(4):4225, 1995.

\bibitem{Sa09}
David Saad.
\newblock {\em On-line learning in neural networks}, volume~17.
\newblock Cambridge University Press, 2009.

\bibitem{GASKZ19}
Sebastian Goldt, Madhu Advani, Andrew~M Saxe, Florent Krzakala, and Lenka
  Zdeborov{\'a}.
\newblock Dynamics of stochastic gradient descent for two-layer neural networks
  in the teacher-student setup.
\newblock In {\em Advances in Neural Information Processing Systems}, pages
  6979--6989, 2019.

\bibitem{goldt2019modelling}
Sebastian Goldt, Marc M{\'e}zard, Florent Krzakala, and Lenka Zdeborov{\'a}.
\newblock Modelling the influence of data structure on learning in neural
  networks.
\newblock {\em arXiv preprint arXiv:1909.11500}, 2019.

\bibitem{rotskoff2018neural}
Grant~M Rotskoff and Eric Vanden-Eijnden.
\newblock Neural networks as interacting particle systems: Asymptotic convexity
  of the loss landscape and universal scaling of the approximation error.
\newblock {\em arXiv preprint arXiv:1805.00915}, 2018.

\bibitem{MMN18}
Song Mei, Andrea Montanari, and Phan-Minh Nguyen.
\newblock A mean field view of the landscape of two-layer neural networks.
\newblock {\em Proceedings of the National Academy of Sciences},
  115(33):E7665--E7671, 2018.

\bibitem{chizat2018global}
Lenaic Chizat and Francis Bach.
\newblock On the global convergence of gradient descent for over-parameterized
  models using optimal transport.
\newblock In {\em Advances in neural information processing systems}, pages
  3036--3046, 2018.

\bibitem{NIPS2016_6322}
Ben Poole, Subhaneil Lahiri, Maithra Raghu, Jascha Sohl-Dickstein, and Surya
  Ganguli.
\newblock Exponential expressivity in deep neural networks through transient
  chaos.
\newblock In D.~D. Lee, M.~Sugiyama, U.~V. Luxburg, I.~Guyon, and R.~Garnett,
  editors, {\em Advances in Neural Information Processing Systems 29}, pages
  3360--3368. Curran Associates, Inc., 2016.

\bibitem{schoenholz2017deep}
Samuel~S. Schoenholz, Justin Gilmer, Surya Ganguli, and Jascha Sohl-Dickstein.
\newblock Deep information propagation, 2017.

\bibitem{yang2018a}
Greg Yang, Jeffrey Pennington, Vinay Rao, Jascha Sohl-Dickstein, and Samuel~S.
  Schoenholz.
\newblock A mean field theory of batch normalization.
\newblock In {\em International Conference on Learning Representations}, 2019.

\bibitem{mei2019meanfield}
Song Mei, Theodor Misiakiewicz, and Andrea Montanari.
\newblock Mean-field theory of two-layers neural networks: dimension-free
  bounds and kernel limit, 2019.

\bibitem{gilboa2019dynamical}
Dar Gilboa, Bo~Chang, Minmin Chen, Greg Yang, Samuel~S. Schoenholz, Ed~H. Chi,
  and Jeffrey Pennington.
\newblock Dynamical isometry and a mean field theory of lstms and grus, 2019.

\bibitem{novak2019bayesian}
Roman Novak, Lechao Xiao, Yasaman Bahri, Jaehoon Lee, Greg Yang, Daniel~A.
  Abolafia, Jeffrey Pennington, and Jascha Sohl-dickstein.
\newblock Bayesian deep convolutional networks with many channels are gaussian
  processes.
\newblock In {\em International Conference on Learning Representations}, 2019.

\bibitem{MPV87}
Marc M\'ezard, Giorgio Parisi, and Miguel~A. Virasoro.
\newblock {\em Spin glass theory and beyond}.
\newblock World Scientific, Singapore, 1987.

\bibitem{GKKR96}
Antoine Georges, Gabriel Kotliar, Werner Krauth, and Marcelo~J Rozenberg.
\newblock Dynamical mean-field theory of strongly correlated fermion systems
  and the limit of infinite dimensions.
\newblock {\em Reviews of Modern Physics}, 68(1):13, 1996.

\bibitem{PUZ20}
Giorgio Parisi, Pierfrancesco Urbani, and Francesco Zamponi.
\newblock {\em Theory of Simple Glasses: Exact Solutions in Infinite
  Dimensions}.
\newblock Cambridge University Press, 2020.

\bibitem{Gabri__2020}
Marylou Gabri{\'{e}}.
\newblock Mean-field inference methods for neural networks.
\newblock {\em Journal of Physics A: Mathematical and Theoretical},
  53(22):223002, may 2020.

\bibitem{arous1997symmetric}
G~Ben Arous, Alice Guionnet, et~al.
\newblock Symmetric langevin spin glass dynamics.
\newblock {\em The Annals of Probability}, 25(3):1367--1422, 1997.

\bibitem{SBCKUZ20}
Stefano~Sarao Mannelli, Giulio Biroli, Chiara Cammarota, Florent Krzakala,
  Pierfrancesco Urbani, and Lenka Zdeborov{\'a}.
\newblock Marvels and pitfalls of the langevin algorithm in noisy
  high-dimensional inference.
\newblock {\em Physical Review X}, 10(1):011057, 2020.

\bibitem{SKUZ19}
Stefano~Sarao Mannelli, Florent Krzakala, Pierfrancesco Urbani, and Lenka
  Zdeborova.
\newblock Passed \& spurious: Descent algorithms and local minima in spiked
  matrix-tensor models.
\newblock In {\em international conference on machine learning}, pages
  4333--4342, 2019.

\bibitem{FPSUZ17}
Silvio Franz, Giorgio Parisi, Maxim Sevelev, Pierfrancesco Urbani, and
  Francesco Zamponi.
\newblock Universality of the sat-unsat (jamming) threshold in non-convex
  continuous constraint satisfaction problems.
\newblock {\em SciPost Physics}, 2(3):019, 2017.

\bibitem{FHU19}
Silvio Franz, Sungmin Hwang, and Pierfrancesco Urbani.
\newblock Jamming in multilayer supervised learning models.
\newblock {\em Physical review letters}, 123(16):160602, 2019.

\bibitem{mignacco2020role}
Francesca Mignacco, Florent Krzakala, Yue~M Lu, and Lenka Zdeborov{\'a}.
\newblock The role of regularization in classification of high-dimensional
  noisy gaussian mixture.
\newblock {\em arXiv preprint arXiv:2002.11544}, 2020.

\bibitem{rosset2004margin}
Saharon Rosset, Ji~Zhu, and Trevor~J Hastie.
\newblock Margin maximizing loss functions.
\newblock In {\em Advances in neural information processing systems}, pages
  1237--1244, 2004.

\bibitem{deng2019model}
Zeyu Deng, Abla Kammoun, and Christos Thrampoulidis.
\newblock A model of double descent for high-dimensional binary linear
  classification.
\newblock {\em arXiv preprint arXiv:1911.05822}, 2019.

\bibitem{SOS92}
H~Sebastian Seung, Manfred Opper, and Haim Sompolinsky.
\newblock Query by committee.
\newblock In {\em Proceedings of the fifth annual workshop on Computational
  learning theory}, pages 287--294, 1992.

\bibitem{ABUZ18}
Elisabeth Agoritsas, Giulio Biroli, Pierfrancesco Urbani, and Francesco
  Zamponi.
\newblock Out-of-equilibrium dynamical mean-field equations for the perceptron
  model.
\newblock {\em Journal of Physics A: Mathematical and Theoretical},
  51(8):085002, 2018.

\bibitem{De76}
Cirano~de Dominicis.
\newblock Technics of field renormalization and dynamics of critical phenomena.
\newblock In {\em J. Phys.(Paris), Colloq}, pages C1--247, 1976.

\bibitem{EO92}
H~Eissfeller and M~Opper.
\newblock New method for studying the dynamics of disordered spin systems
  without finite-size effects.
\newblock {\em Physical review letters}, 68(13):2094, 1992.

\bibitem{SHNGS18}
Daniel Soudry, Elad Hoffer, Mor~Shpigel Nacson, Suriya Gunasekar, and Nathan
  Srebro.
\newblock The implicit bias of gradient descent on separable data.
\newblock {\em The Journal of Machine Learning Research}, 19(1):2822--2878,
  2018.

\bibitem{Ku02}
Jorge Kurchan.
\newblock Supersymmetry, replica and dynamic treatments of disordered systems:
  a parallel presentation.
\newblock {\em arXiv preprint cond-mat/0209399}, 2002.

\bibitem{Ku92}
Jorge Kurchan.
\newblock Supersymmetry in spin glass dynamics.
\newblock {\em Journal de Physique I}, 2(7):1333--1352, 1992.

\bibitem{Zi96}
Jean Zinn-Justin.
\newblock {\em Quantum field theory and critical phenomena}.
\newblock Clarendon Press, 1996.

\bibitem{EO94}
H~Eissfeller and M~Opper.
\newblock Mean-field monte carlo approach to the sherrington-kirkpatrick model
  with asymmetric couplings.
\newblock {\em Physical Review E}, 50(2):709, 1994.

\bibitem{RBBC19}
Felix Roy, Giulio Biroli, Guy Bunin, and Chiara Cammarota.
\newblock Numerical implementation of dynamical mean field theory for
  disordered systems: application to the lotka--volterra model of ecosystems.
\newblock {\em Journal of Physics A: Mathematical and Theoretical},
  52(48):484001, 2019.

\bibitem{MSZ20}
Alessandro Manacorda, Gr{\'e}gory Schehr, and Francesco Zamponi.
\newblock Numerical solution of the dynamical mean field theory of
  infinite-dimensional equilibrium liquids.
\newblock {\em The Journal of Chemical Physics}, 152(16):164506, 2020.

\end{thebibliography}

\end{document}